\documentclass[10pt,journal,compsoc]{IEEEtran}
\ifCLASSOPTIONcompsoc
\usepackage[nocompress]{cite}
\else
\usepackage{cite}
\fi
\usepackage{amssymb}
\usepackage{times}
\usepackage{latexsym}
\usepackage{verbatim}

\usepackage{microtype}

\usepackage{multirow}
\usepackage{booktabs}
\usepackage{bigstrut}
\usepackage{hyperref}
\usepackage{makecell}
\usepackage{pifont}
\usepackage{amssymb}
\usepackage{bbding}
\usepackage{float}
\usepackage{graphicx}
\usepackage{subcaption}
\usepackage{amsmath,bm}
\usepackage{amsmath}
\usepackage{color}
\usepackage[ruled]{algorithm}
\usepackage{algorithmic}
\usepackage{setspace}
\usepackage{bbding}

\makeatletter

\newcommand{\Rmnum}[1]{\expandafter\@slowromancap\romannumeral #1@}
\newcommand{\tabincell}[2]{\begin{tabular}{@{}#1@{}}#2\end{tabular}}
\makeatother

\usepackage{microtype}

\begin{document}

	\title{Multi-Scale Feature and Metric Learning for Relation Extraction}

	\author{Mi Zhang,~\IEEEmembership{Student Member,~IEEE,} {Tieyun~Qian,~\IEEEmembership{Member,~IEEE} 
			\IEEEcompsocitemizethanks{\IEEEcompsocthanksitem Mi Zhang and Tieyun Qian are with Wuhan University, China. \protect E-mail: mizhanggd@whu.edu.cn, qty@whu.edu.cn, Corresponding author: Tieyun Qian.}
		}
		\thanks{Manuscript received July 28, 2021; revised,.}}

	\markboth{Journal of \LaTeX\ Class Files,~Vol.~14, No.~8, August~2015}%
	{Shell \MakeLowercase{\textit{et al.}}: Bare Demo of IEEEtran.cls for Computer Society Journals}
	
	\IEEEtitleabstractindextext{%
		\begin{abstract}
			Existing methods in relation extraction have leveraged the lexical features in the word sequence and the syntactic features in the parse tree. Though effective, the lexical features extracted from the successive word sequence may introduce some noise that has little or no meaningful content. Meanwhile, the syntactic features are usually encoded via graph convolutional networks  which  have restricted receptive field.
			
			To address the above limitations, we propose a multi-scale feature and metric learning framework for relation extraction. Specifically, we first develop \emph{a multi-scale convolutional neural network} to aggregate the non-successive mainstays in the lexical sequence. We also design \emph{a  multi-scale graph convolutional network} which can increase the receptive field towards specific syntactic roles. Moreover, we present \emph{a multi-scale metric learning} paradigm to exploit both the feature-level relation between lexical and syntactic features and the sample-level relation between instances with the same or different classes.
			We conduct extensive experiments on three real world datasets for various types of relation extraction tasks. The results demonstrate that our model significantly outperforms the state-of-the-art approaches.
		\end{abstract}
		
		}

	\maketitle
	\IEEEdisplaynontitleabstractindextext
	\IEEEpeerreviewmaketitle

	\section{Introduction}
	The goal of relation extraction (RE) is to predict the semantic relations among entities in the text. RE plays a vital role in many applications such as question answering~\cite{YuYHSXZ17,LuanHOH18,BhutaniZJ19}, knowledge graph population~\cite{ZhangZCAM17,BosselutRSMCC19,SaRR0W0Z17}, and recommender systems~\cite{BelliniANS17,HuangFQSLX19}. As a result, it has aroused tremendous research interests in recent years.
	
\begin{figure}[htb]
		\center{\includegraphics[width=0.42\textwidth]{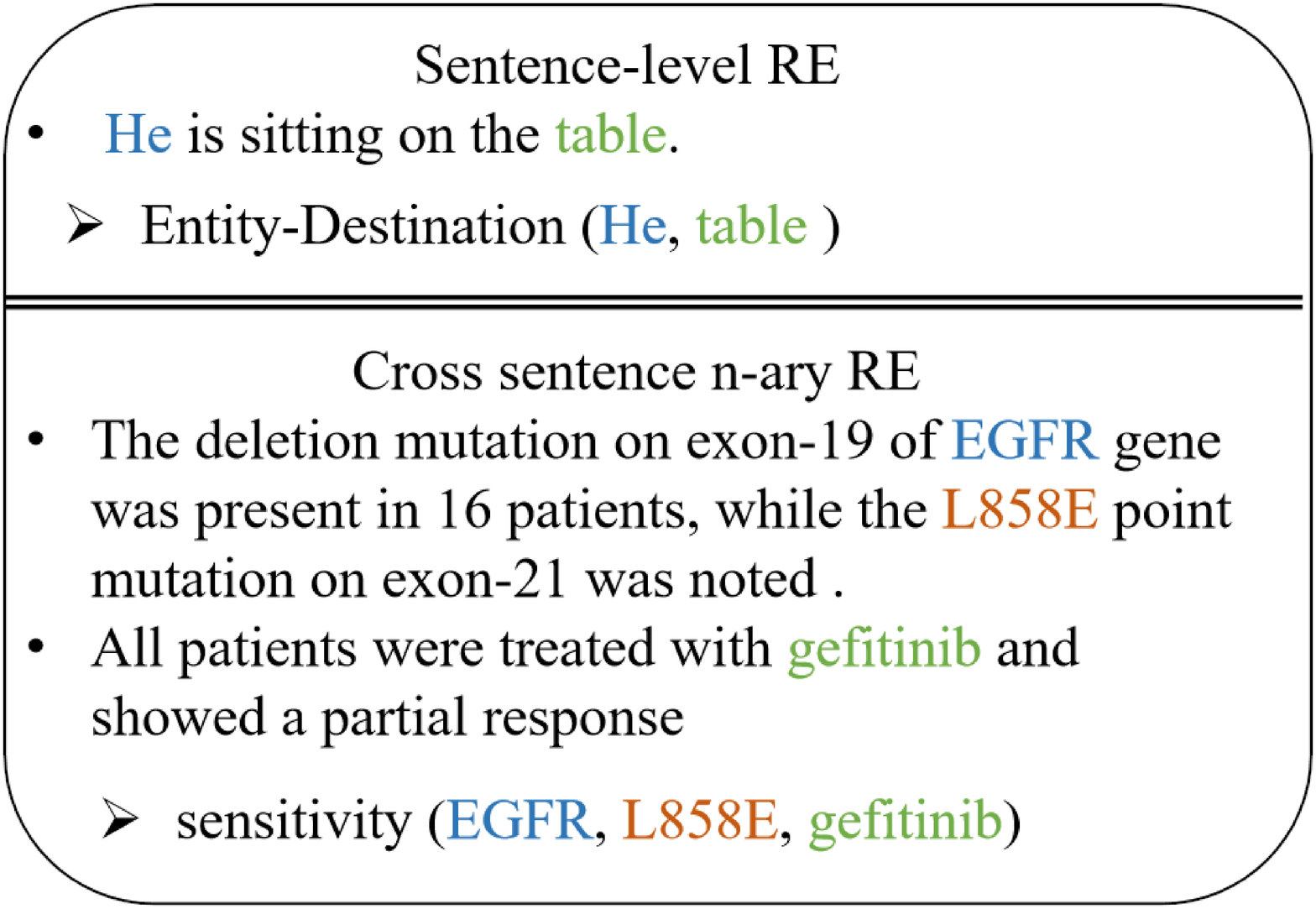}}
		\caption{ Examples of  a binary relation between two entities in one sentence from the Semeval dataset and a ternary relation among three entities across two sentences from the PubMed dataset.}
		\label{fig:instance}
		\vspace{-5mm}
	\end{figure}

	There are mainly two types of RE tasks, namely, the binary relation extraction in single sentences~\cite{ZelenkoAR02,HasegawaSG04,Bach2007ARO} and  the cross-sentence $n$-ary relation extraction~\cite{mcdonald-etal-2005-simple,GerberC10,YoshikawaRHAM10}. Figure~\ref{fig:instance} shows two examples of sentence-level relation and cross-sentence $n$-ary relation extraction.
	For sentence-level RE, given two entities `He' and `table', we aim to extract the gold relation `Entity-Destination'.
	For cross-sentence $n$-ary RE, given three target entities `EGFR', `L858E', `gefitinib' in a piece of text consisting of two sentences, we should extract the gold ternary relation `sensitivity'. Note two sentences convey the fact that there is a ternary interaction among  three entities, which is not expressed in either sentence alone.
	
	Current studies in RE focus on feature learning that discovers the representations from raw data. The word sequence has abundant lexical information and the dependency graph of the sentence carries unique syntactic information and they can both make contributions to RE tasks.
	Consequently, existing RE models can be roughly categorized into two classes: sequence-based and dependency-based. Sequence-based models~\cite{SocherHMN12,LiuSCC13,ZengLLZZ14,WangCML16,MiwaB16,TourilleFNT17} extract local and global lexical features from the word sequence  using  convolutional neural network (CNN)~\cite{Kim14} and long short-term memory (LSTM)~\cite{HochreiterS97}. Dependency based models operate on parse trees either to extract paths between entities as features or as kernels in a statistical classifier or to construct tree or graph structure, such that tree- (graph-) LSTM or graph convolution network (GCN) can be applied to getting the representations of all words~\cite{BunescuM05,SongZWG18,GuoZL19}, where GCNs  are widely adopted in RE tasks due to their superior performance.
	There are  also a few  methods~\cite{Zhang0M18,GuoZL19,MandyaBC20,LeeSOLSL20} working in a hybrid mode with both the lexical and syntactic features.
	All the above methods  employ the static word embeddings as the input. Recently, several studies~\cite{SoaresFLK19,WuH19a,YamadaASTM20,PengGHLLLSZ20} adopt the pre-trained language models (PLMs) like BERT~\cite{DevlinCLT19} or RoBERTa~\cite{Liu2020Myle} to get the contextualized word embeddings which can further improve the performance.
	

	Despite of the effectiveness, existing sequence-based and dependency-based models have inherent limitations.
	\begin{itemize}
		\item Convolutional neural networks (CNNs)~\cite{Kim14} are the typical architecture used to extract local information from word sequences~\cite{ZengLLZZ14,WangCML16}, which capture successive lexical features in a word by word manner. This will introduce some noise since languages often contain a large number of tokens that have little or no meaningful content. For example, give the sentence `He is sitting on the table', the auxiliary verb `is' and the determiner `the' do not have concrete meanings.
		\item Graph convolution networks (GCNs) ~\cite{Zhang0M18,GuoZL19} over parse trees only have restricted receptive field along the dependency path. In the aforementioned example, the goal is to extract the `Entity-Destination' relation `on' between two entities `He' and `table'.  However, the node `He' cannot be aggregated with  the node `on' via a standard convolution operation since they are located in different paths, and simply adding convolutional layers means performing more Laplacian smoothing~\cite{LiHW18_Deeper} and will make the output features similar.
	\end{itemize}

	To tackle the above limitations, we propose a multi-scale feature and metric learning framework for relation extraction.
	We first design a multi-scale feature learning  module  to encode the multi-scale lexical and syntactic information in the word sequence and parse tree. To be specific, we present a multi-scale convolutional neural network (\emph{MSCNN}) for extracting the mainstays in the lexical sequence. MSCNN contains \emph{a horizontal dilated convolution layer} and \emph{a vertical convolution layer}. The dilated horizontal convolution enables the model focus on the important yet non-successive words in the sentence  like `He' and `sitting'. The vertical convolution captures the dimension-level sequential patterns which can aggregate every dimension in other words' latent representations.
	
	We also develop a multi-scale graph convolution networks (\emph{MSGCN}) to obtain \emph{the coarsened and refined view of the parse tree}. Instead of merely aggregating one- or two-hop neighborhood syntactic information, MSGCN enlarges the receptive field for the word via the construction of coarsened syntactic graph, and thus more local syntactic information can be closely related. For example, `He' and `table' are merged into a hypernode in the coarsened graph where the `He' node can see the `table' node's neighboring words. As a result,  the neighbors of `table' like `on' can be aggregated with `He' in one convolution operation.
	
	We finally introduce a multi-scale metric learning (\emph{MSML}) paradigm to exploit both \emph{the feature-level and relation-level relations}, such that we can pull the neighbors in the feature and relation space close and push  those non-neighbors apart. On one hand, both the syntactic and lexical features of a sentence refer to the same word sequence. Though their representations might be different, their relative distance in the feature space should be smaller than that of this lexical embedding to the syntactic embedding of another sentence. On the other hand, for the RE task, the same/different relations should be located nearby/apart from each other in the relation space, respectively.

	Our model is among the first to exploit the  hierarchical nature in lexical, syntactic, and relational representations of the text data for RE tasks. To validate our proposed model, we conduct extensive experiments on three popular datasets for the sentence-level and cross sentence n-ary RE tasks. With either the static  word embedding GloVe~\cite{PenningtonSM14} or the contextualized word embedding BERT and RoBERTa~\cite{DevlinCLT19,Liu2020Myle}, our model outperforms the state-of-the-art counterpart methods for both tasks on all three datasets.

	\section{Related Work}
	We briefly review the literature in relation extraction,  multi-scale representation learning, and metric learning.
	
	\subsection{Relation Extraction Models}
	According to the features used in the models, most existing relation extraction methods can be categorized into  lexicon based (aka sequence based) and syntax based (aka dependency based) ones.
	Early statistical RE models use the relational structure of the verbs to discriminate relations ~\cite{AoneR00,TemkinG03}, or employ similarity of context word as the features or kernels ~\cite{ChenJTN05a,HasegawaSG04,BunescuM05}. Recently, several types of neural models, e.g., CNN based~\cite{LiuSCC13,ZengLLZZ14,NguyenG15,LinSLLS16}, RNN based~\cite{ZhangW15a,LinMDABS18}, and LSTM based~\cite{ZhouSTQLHX16,TourilleFNT17}, have been proposed to extract better lexical features from word sequences.
	
	The syntactic information  is also effective in improving relation extraction performance as it can capture long-distance relations. Various models have been developed to incorporate syntactic information. For example, LSTM and entity position-aware attention ~\cite{ZhangZCAM17} using POS tags, tree or graph structure  are employed as part of word embeddings. Tree or graph structures \cite{XuMLCPJ15,MiwaB16,Zhang0M18,GuoZL19,LeeSOLSL20} are used to encode dependency relations as embeddings where GCN becomes the mainstream due to its superior performance. Several methods ~\cite{MiwaB16,ZhangZCAM17,PengPQTY17,SongZWG18} pay attention to the pruning strategies for selecting relevant partial dependency structures. More recently, dependency forest \cite{GuoN0C20} or multiple dependency sub-graphs  \cite{MandyaBC20} are constructed to  obtain more informative features.
	
	Besides the above sequence and dependency based methods which adopt the static word embeddings as input, several recent studies employ the contextualized word embedding like BERT or RoBERTa for RE tasks based on the fine-tuning or post-training techniques. For example,  LUKE~\cite{YamadaASTM20} uses an improved transformer architecture with an entity-aware self-attention mechanism to enhance entity-related representations. R-BERT~\cite{WuH19a}  incorporates information from the target entities to tackle the relation classification task. CP~\cite{PengGHLLLSZ20} presents the entity-masked contrastive post-training framework for RE using external knowledge base.
	

	Overall, existing methods either capture information word by word sequentially or do not fully exploit  the dependency along different syntactic paths.
	The sequence based methods extract successive information from the word sequence and thus may introduce some noise. Meanwhile, the dependency based GCN methods can only convolute along the syntactic paths. In contrast, we present the multi-scale CNN and multi-scale GCN which capture the non-successive information in the sentence and expand the receptive field for the nodes in the parse tree. Also note that the utilization of contextualized word embedding is not the focus of our paper, hence we design a simple post-training method and integrate its output into the  input embedding layer of our model.

	\subsection{Multi-Scale Representation Learning}
	Multi-scale information is of great significance in computer vision area. For example, FPN~\cite{LinDGHHB17} sends images of different scales to the network to extract the features of different scales for integration. D-CNN~\cite{YuK15} develops a convolutional network module that aggregates multi-scale contextual information without losing resolution or analyzing re-scaled images.
	Some work has also been proposed for learning multi-scale information from graphs. For example, Chen et al. \cite{ChenPHS18} propose a coarsening procedure to construct a coarsened graph of smaller size and use graph network to encode graph node embedding, and they then use a refining procedure to get the original graph embedding. HGCN~\cite{Hu0WWT19} repeatedly aggregates structurally similar nodes into hypernodes and then refines the coarsened graph to the original one.
	
	Inspired by these studies, we introduce the multi-scale learning technique into the natural language processing field, and develop a novel multi-scale feature and relation learning paradigm tailored for RE tasks.

	\subsection{Metric Learning}
	Metric learning~\cite{XingNJR02} produces distance metrics that capture important relationships among data and is an indispensable technique for many successful machine learning applications, including image classification, document retrieval, protein function prediction, and recommender systems~\cite{ChenKDD12,TaigmanYRW14,WanWHWZZL14,KostingerHWRB12,XuCWS12}.
	
	The object of metric learning is to reduce or limit the distance between similar pairs and increase the distance between dissimilar pairs through the training and learning stage. The global optimization essentially attempts to learn a distance metric that pulls all similar pairs together, and pushes dissimilar pairs apart. Metric learning is usually applied to individual-level fine-grained face recognition ~\cite{SchroffKP15}. In addition to the computer vision, metric learning is also adopted in many recommendation algorithms based on the triplet loss, where  the items in the candidate set may all be liked by the user, and the triplet loss is used to compare which product is more suitable for the user~\cite{HsiehYCLBE17}.
	
	We apply metric learning to RE tasks, and we extend it into a multi-scale version that includes the feature level and relation level metric learning. By doing this, we can fully explored the neighborhood information in both the feature and relation space.

	\section{Preliminary}
	\label{sec:length}
	\subsection{Task Definition}
	Relation extraction tasks are mainly categorized into two types: sentence-level and cross-sentence n-ary RE~\footnote{There are several studies towards document-level RE where the relations are expressed across multiple paragraphs, and extra efforts are required to deal with the multiple mentions of an entity. Its learning focus  is different from that of sentence-level and cross-sentence n-ary RE and thus we omit it.}. Formally,
	let \textit{T} = [$t_{1}$, ..., $t_{m}$] be a text consisting of sentence(s) with $m$ tokens (words), and \textit{e$_1$}, ..., \textit{e$_n$} be the entity mentions in \textit{T}. $R$ = [$r_{1}$, ..., $r_{o-1}$, $r_{o}$]~($r_{o}$ = None) is a predefined relation set.
	The relation extraction task can be formulated as a classification problem of determining whether a relation $r$ $\in$ $R$ holds for \textit{e$_1$}, ..., \textit{e$_n$}. More specifically, sentence-level RE focuses on the binary relations where two entities occur in the same sentence (i.e., $n$ = 2 and \textit{T} is a sentence) while cross-sentence n-ary RE focuses on multiple sentences with entities (i.e., $n > $  2 and \textit{T} contains multiple sentences).
	
	\subsection{Anchor Triplet}
	We sample a positive and negative sample for an anchor to form a triplet before training our model, i.e.,  the input of our model is the anchor triplet ($S$, $S^+$, $S^-$):
	
	$S$: the anchor sample, which is currently going to be classified;
	
	$S^+$: the positive sample having the same class label with $S$;
	
	$S^-$: the negative sample having the different class label with $S$.
	
	Each sample in the anchor triplet contains a word sequence and a dependency graph of an instance in training set. The anchor triplet will be used for metric learning.  Our goal of metric learning is that the trained model can reduce the distance between samples from the same class ($S$, $S^+$) and increase the distance between the different classes ($S$, $S^-$).

	\subsection{Preparing Input Representation}
	For a thorough comparison with various methods, we have prepared two types of word embeddings as the input of our model. One is the static GloVe embedding~\cite{PenningtonSM14} that is widely used in sequence based and dependency based methods. The other is the contextualized embedding including BERT~\cite{DevlinCLT19} and RoBERTa~\cite{Liu2020Myle} used in  pre-training language model (PLM) based methods where the word embeddings are further refined via fine-tuning or post-training in the downstream tasks.

	\subsubsection{Static word embedding}
	Let \textbf{E$_{w}$} $\in$ $\mathbb{R}^{\vert V\vert \times da}$ be the static word embedding~\cite{PenningtonSM14}  lookup table, where $\vert V\vert$ is the vocabulary size and $da$ is the dimension of word embedding. \textbf{E$_{w}$} is used to map the text \textit{T} to a list of word vectors [$\bm{t}{_1^w}$, ..., $\bm{t}{_m^w}$] $\in$ $\mathbb{R}$$^{m\times da}$. Let \textbf{E$_{o}$} $\in$ $\mathbb{R}$$^{m\times db}$ and \textbf{E$_{pos}$} $\in$ $\mathbb{R}$$^{m\times dc}$  be the position and POS embedding lookup table with random initialization, where $db$ and $dc$ are the dimension of position and POS embedding, respectively. \textbf{E$_{o}$} and \textbf{E$_{pos}$} are used to map the the position of tokens and the POS tags of tokens to embeddings.
	
	Overall, each token $t_{i}$ in \textit{T} is vectorized as $\bm{t}{_i^{\textit{wop}}}$ = $\bm{t}{_i^{w}}$ $\oplus$ $\bm{t}{_i^{o}}$ $\oplus$ $\bm{t}{_i^{pos}}$ $\in$ $\mathbb{R}^{{da+n \times db+dc}}$,  where $\bm{t}{_i^{\textit{wop}}}$ contains word, position, and POS information. We use $\bm{t}{^{\textit{wop}}} = [\bm{t}{_1^{\textit{wop}}},...,\bm{t}{_m^{\textit{wop}}}]$ as the input for the token in word sequence and dependency graph.

	\begin{figure*}[htb]
		\center{\includegraphics[width=0.88\textwidth]{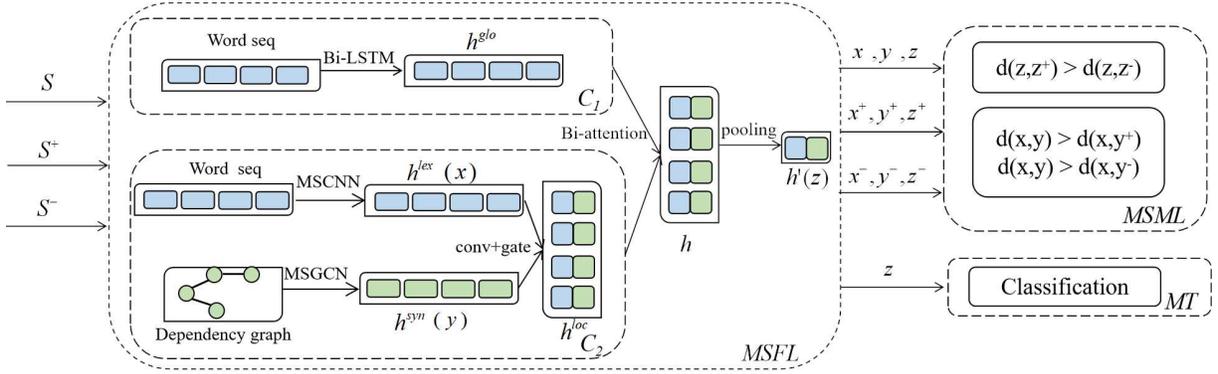}}
		\caption{Architecture of our model}
		\label{fig:model}
		\vspace{-5mm}
	\end{figure*}
	
	\subsubsection{Contextualized word embedding}
	Generally, the methods with contextualized embeddings achieve significantly better performance over those with static embeddings. However, this can be largely due to the power of PLMs rather than the methods themselves. In view of this, we design a simple post-training task on RE training set to obtain contextualized word embeddings for RE tasks. Our method is adapted from previous models with slight modification. This ensures that we can pay more attention to the design of multi-scale feature and metric learning. Below we present the detail.
	
	Firstly, to  make PLMs capture the location information of entities, we follow the method in ~\cite{WuH19a,SoaresFLK19} by inserting a special token `[E1]' and `[E2]' at both the beginning and the end of the first and the second entity, respectively. We also add `[CLS]' to the beginning of sentences. For example, after inserting the special separate tokens, the running sample sentence with target entities `He' and `table' is transformed into:
	
	`\textit{[CLS] [E1] He [E1] is sitting on the [E2] table [E2].}
	
	For each instance in the dataset, we feed the transformed token sequence into PLMs including BERT~\cite{DevlinCLT19} and RoBERTa~\cite{Liu2020Myle}  and obtain the hidden state output of each token.
	
	Secondly, we perform the post-training task on RE training set, which consists of an entity type prediction task and a relation classification task.
	Each entity has its own entity type information. For example, in the PubMed dataset, there are three entity types: grug, gene, and mutation. We  predict the entity type for each entity with its own embedding and the special token embedding `[E1]' and `[E2]'. Specifically, we apply the average operation to get the entity vector representation for each of the entities, and we then use this entity vector to predict its entity type.
	For the post-training relation classification task, we concatenate each entity's hidden embedding and the final hidden state vector of the first token `[CLS]', and then add a fully connected layer and a softmax layer for classification. Besides, the predicted entity type embeddings are also added to the entity embeddings to provide more information for the entities.

	After post-training with these two tasks, the output token embeddings will serve as the inputs for the tokens in word sequence and dependency graph.

	\section{Proposed Model}
	\subsection{Model Overview}
	In this section, we give an overview of our model. We show its architecture in Figure\ref{fig:model}. As can be seen, our model is composed of three main modules: \textit{MSFL}, \textit{MSML}, and \textit{MT}.
	
	\textit{MSFL} is a multi-scale feature learning module which contains two components, denoted as \textit{C}$_1$ and \textit{C}$_2$ in two dotted rectangles. It takes the triplet [$S$, $S^+$, $S^-$] as the input and output three types of representations that used for multi-scale metric learning (\textit{MSML}) and the main classification task (\textit{MT}).
	\textit{C}$_1$ adopts a normal Bi-LSTM to get global feature of the sample.  \textit{C}$_2$ applies the proposed MSCNN  and MSGCN to sequence and dependency graph to capture multi-scale local features.
	
	\textit{MSML} is used for multi-scale metric learning, including the feature level ML that enhances/lessens the relation between the lexical and syntactic features of the same/different sentence and lessens that of the different sentences,  and the relation level ML which strengthens/weakens the relation representation of the same/different classes.
	
	\textit{MT} performs the main relation classification task based on the combined joint representation from \textit{MSFL}.

	\subsection{Multi-Scale Feature Learning (MSFL) Module}
	The word sequence has abundant lexical information and the dependency graph carries unique syntactic information, both of which can make contributions to RE tasks.  CNNs and GCNs have shown effectiveness  in extracting lexical and dependency information. However, as we analyze them in the introduction part, CNNs tend to extract successive local features and GCNs have limited receptive fields. In order to tackle these problems, we propose in this section  our \textit{MSFL} module, which consists of the multi-scale convolution neural network (MSCNN) and multi-scale graph convolution network (MSGCN) beyond the commonly used CNNs and GCNs.
	

	\subsubsection{Multi-scale lexical feature learning}
\begin{figure}[htb]
		\center{\includegraphics[width=0.5\textwidth]{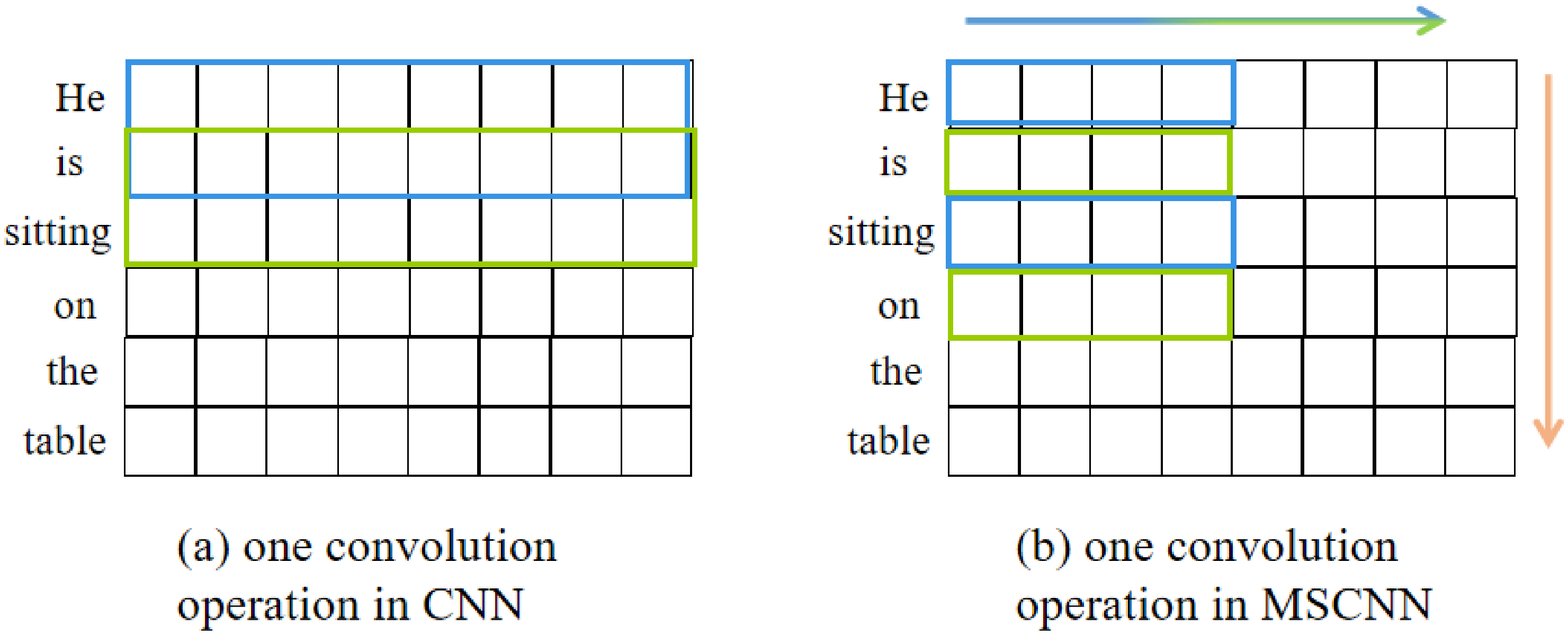}}
		\caption{A sample of original CNN and our proposed MSCNN.}
		\label{fig:mscnn}
		\vspace{-2mm}
	\end{figure}

	The core idea in CNNs is to capture local features. A convolution operation involves a filter which is applied to a  sliding window composed of several words  to produce a new feature. As shown in Fig.~\ref{fig:mscnn} (a), for an ordinary CNN, when the filter size is 2, the window slides over the sequence and convolutes the successive 2-gram such as  `He-is' in blue rectangle and `is-sitting' in green rectangle at each step. However, the word `is' is an auxiliary verb which does not have concrete meaning. Indeed, it is better to put `He' and `sitting' in one window given the fixed filter size 2.  To this end, we apply the dilated convolution method D-CNN~\cite{YuK15} to our RE task, and extend it by convoluting along \emph{both the vertical and horizontal directions} to systematically aggregate lexical information. We term this two-dimensional convolution as MSCNN.
	
	
	
	The two-dimensional convolution on sentences is performed with three different  convolution kernels (kernel size = [3,4,5]) on word-dimension (horizontal direction) and one convolution kernel (kernel size = d/2, where d is the dimension of input representation) on vector-dimension (vertical direction). On one hand, the horizontal dilated convolution enables the non-successive tokens like `He' and `sitting' to be convoluted in one sliding window. On the other hand, the vertical convolution captures the dimension-level sequential patterns such that the corresponding dimensions of the words in the same window can be aggregated. The illustration of MSCNN is shown in in Fig.~\ref{fig:mscnn} (b), where the horizontal convolution has a dilation rate 2 (the rectangles in the same color are in one window), and the green and orange  arrows denote the horizontal and vertical direction convolutions, respectively. We do not show the vertical convolution for clarity.

	Formally, a feature $c_i$ is generated with two-dimensional convolution filter $w_c$ for a token window $\bm{x}_{i:i+j-1}$. By sliding the filter window from the beginning of the text \textit{T} till the end, we get a feature map $\bm{c}$. After that, we use a max pooling $f_3(\cdot)$ to get one single general text feature:
	\begin{equation}
	\vspace{-2mm}
	\small
	c_i = \sum_{k=1}^{K} \bm{t}{^{\textit{wop}}}[i + rk]w[k]
	\end{equation}
	
	\begin{equation}
	\vspace{-1mm}
	\small
	\bm{c} = [c_1, c_2, ..., c_{m-j+1}],~
	\bm{h}^{\textit{lex}}_i = f(\bm{c}),
	\end{equation}

	where $\bm{t}{^{\textit{wop}}}$ is the input, $i$ is the input index, $k$ is the index in the convolution kernel, $r$ is dilated rate ($r$ = 2), $\odot$ represents element-wise product, and $f({\cdot})$ is a non-linear function.
	The vertical convolution can be defined in a similar.
	After performing three different horizontal convolution kernels with the dilation factor $r$ and one vertical convolution kernel, we can obtain the refined multi-scale lexical feature embedding $\bm{x}$ for the text $T$,
	\begin{equation}
	\vspace{-1mm}
	\small
	\bm{x} = \bm{h}^{\textit{lex}} = [\bm{h}{_{1}^{\textit{lex}}}, ..., \bm{h}{_{m}^{\textit{lex}}}]
	\end{equation}

	\subsubsection{Multi-scale syntactic feature learning}
\begin{figure}[htb]
		\center{\includegraphics[width=0.5\textwidth]{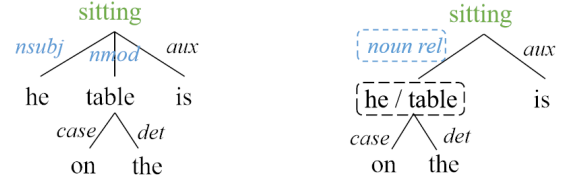}}
		\caption{A sample of the original syntactic graph in GCN and the coarsened syntactic graph in our MSGCN.}
		\label{fig:case}
		\vspace{-2mm}
	\end{figure}

	GCNs are based on the neighborhood aggregation scheme which combines information from neighborhoods to generate node embedding. In our task, one single  GCN operation acts as the approximation of aggregation on the first-order neighbors along the dependency path. As shown in Figure~\ref{fig:case} (a), `he', `table', and `is' can be aggregated  to the `sitting' node, whereas `on' is the three-hop neighbor of `he' and thus cannot be aggregated into  `he' in one or two graph convolution operations. Meanwhile, adding too many aggregations will cause the nodes' features over-smoothed and make them indistinguishable.
	
	In view of this, we propose a multi-scale graph convolution network (MSGCN). The key idea in our MSGCN is to form a coarsened graph by merging the nodes with similar syntactic roles into a hypernode, and then perform the graph coarsening and refining operation. Fig.~\ref{fig:mscnn} (b) shows an example of the coarsened graph where `he' and `table' are merged into a hypernode.
	

	\textbf{Graph coarsening and refining operation}
	Our proposed MSGCN is inspired by H-GCN~\cite{Hu0WWT19}. However, instead of aggregating structurally similar nodes in  H-GCN~\cite{Hu0WWT19}, we merge the nodes  having the same relation type with a core node into hypernodes to form the coarsened graph. This enables our MSGCN to learn multi-scale syntactic information by enlarging the receptive field for each word in the hypernode. Below we present the detail of graph coarsening operation in MSGCN.
	
	The goal in RE tasks is to extract the relation of the entities, and verbs serve as the predicate of a sentence which express an action, an event, or a state. They are the most important syntactical functions in a sentence. Besides, prepositions show the place, position, time, or method, e.g., `on the table'. Hence we select the entity, verb, and preposition words as \emph{the core node}, and their neighboring words  in the parse tree will be merged into a \emph{hypernode}  if they have the same relation type towards the core word.
	
	
	For example, given the sample `He is sitting on the table', `sitting' is the verb in the sentence and hence is  a core node, and both the nsubj relation `sitting-he' and the nmod relation ` sitting-table' have  the same noun relation type. Consequently, `he' and `table' are grouped into a hypernode `he/tabel', and all their respective neighborhoods are  linked to the hypernode. Also note that  the relation between `he/tabel' and  `sitting' ascends to a relation type of `noun rel'. A new coarsened graph is then formed, as shown in Fig.~\ref{fig:mscnn} (b). Clearly, with this coarsened graph, the entity word `he' has a wide receptive field, and we can obtain the relation `on' between `he' and `table' in one convolution operation  in MSGCN.
	
	
	In order to restore the original topological structure of the graph and retain the original semantic information of words,  we perform the graph refining operation, which is done by splitting the hypernode like `he/tabel' into the original sperate nodes like `he' and `table'.
	
	

	\textbf{Multi-scale GCN}
	MSGCN is built upon graph convolutional networks (GCN)~\cite{KipfW17}. At the  $l^{th}$ layer, GCN takes the graph adjacency matrix $A_l$ and the hidden representation matrix ($\bm{h}_l^{\textit{dep}}$ in our case) as the input, and outputs a hidden representation matrix $\bm{h}_{\textit{l+1}}^{\textit{dep}}$
	, which is described as follows:
	\begin{equation}
	\setlength{\abovedisplayskip}{6pt}
	\setlength{\belowdisplayskip}{6pt}
	\small
	\bm{h}{^{\textit{dep}}_{\textit{l+1}}} = ReLU(\widetilde{D}_l^{-\frac{1}{2}}\widetilde{A}_l \widetilde{D}_l^{-\frac{1}{2}}\bm{h}{_l^{\textit{dep}}}W_l),
	\label{gcn}
	\end{equation}
	where $\bm{h}{^{\textit{dep}}_1} = t^{\textit{wop}}$, $\widetilde{A}_l$ is the adjacency matrix with self-loop $\widetilde{A}_l = A_l + \textit{I}$ , $\widetilde{D}_l $ is the degree matrix of $\widetilde{A}_l$, and $W_l$ is a trainable weight matrix.
	
	In order to learn the multi-scale syntactic features for the word nodes, MSGCN first uses GCN formulated in Eq.~\ref{gcn} to obtain the word node representation, and it then applies GCN to the coarsened and refined graphs, respectively. The embedding of the hypernode in the coarsened graph is the summation of the original nodes' embedding, and the the graph adjacency matrix   is constructed according to the links between the hypernodes and the remaining nodes in the coarsened graph. For the refined graph, the node's embedding is the  the summation of the hypernode's and the original node's embeddings. Since the graph coarsening and refining operation are symmetric, the graph adjacency matrix in the refining graph equals to that in its corresponding  coarsening graph.
	
	By applying the above graph convolution operations to the original, coarsened, and refined graphs, we can finally obtain the multi-scale syntactic feature embedding $\bm{y}$ as follows:
	\begin{equation}
	\vspace{-1mm}
	\small
	\bm{y} = \bm{h}^{\textit{dep}} = [\bm{h}{_{1}^{\textit{dep}}}, ..., \bm{h}{_{m}^{\textit{dep}}}]
	\end{equation}

	\subsubsection{Getting joint representations}
	\textbf{Multi-scale local representations}
	By now we get the multi-scale local representations for word sequence and dependency graph, denoted as $\bm{h}^{\textit{lex}}$($\bm{x}$) and $\bm{h}^{\textit{dep}}$ ($\bm{y}$) in Fig.~\ref{fig:model}. We then employ \textit{conv+gate}~\cite{DauphinFAG17}  mechanism to enrich $\bm{h}^{\textit{dep}}$ with the help of $\bm{h}^{\textit{lex}}$. Specifically, we apply one-layer convolution operation~\cite{Kim14} to $\bm{x}$ with the convolution filter $w_c$.
	We then use the gating mechanism to control the flow of $\bm{x}$ after convolution towards $\bm{y}$, and obtain the enriched multi-scale local representation $\bm{h^{\textit{loc}}}$:
	\begin{equation}
	\vspace{-1mm}
	\setlength{\abovedisplayskip}{6pt}
	\setlength{\belowdisplayskip}{6pt}
	\small
	\bm{g}^{y} = \textit{tanh}(\bm{y} + w_c \ast \bm{x}), ~
	\bm{h^{\textit{loc}}} = \bm{y} \odot \bm{g}^ {y} + \bm{y},
	\label{equ:local}
	\end{equation}
	where $\odot$ represents the element-wise product, and $\ast$ represents one layer convolution.
	
	\textbf{Global representations}
	Since we have got rich local representation from the sentence, we propose to use a Bi-LSTM layer to further obtain global information of the sentence by learning sequential information available in the word sequence. The Bi-LSTM layer takes a series of vectors as inputs, and generates a  hidden state vector in the forward and backward directions for each input.
	
	\begin{equation}
	\setlength{\abovedisplayskip}{6pt}
	\setlength{\belowdisplayskip}{6pt}
	\small
	\bm{h}^{\textit{glo}} = [\bm{h}_{1}^{\textit{glo}}, ..., \bm{h}_{m}^{\textit{glo}}] = \textit{Bi-LSTM}(\bm{t}{_1^{\textit{wop}}}, ..., \bm{t}{_m^{\textit{wop}}}),
	\label{equ:global}
	\end{equation}
	
	\textbf{Joint representations}
	With the above local and global representations $\bm{h^{\textit{glo}}}$ and $\bm{h^{\textit{loc}}}$, we adopt the bi-attention mechanism~\cite{LuYBP16,GaoJYLHWL19} to pass information between $\bm{h}^{\textit{glo}}$ and $\bm{h}^{\textit{loc}}$ according to the attention weights.
	
	Concretely, we first compute the attention score $a_{i,j}$ between the $i^{th}$ global representation and the $j^{th}$ local representation.
	\begin{equation}	
	\setlength{\abovedisplayskip}{6pt}
	\setlength{\belowdisplayskip}{6pt}
	\small
	a_{i,j} = \frac{exp(\bm{h}^{\textit{loc}}_{j} \bm{h}{^{\textit{glo}}_{i}}^{T})} {\sum_{u}^{m} exp(\bm{h}^{\textit{loc}}_{u} \bm{h}{^{\textit{glo}}_{i}}^{T})},
	\end{equation}
	
	We next use $a_{i,j}$ to differentiate each word and get a new word embedding $\bm{\tilde{h}}{^{\textit{glo}}_{i}}$:
	\begin{equation}
	\label{equ:bi-att}
	\small
	\setlength{\abovedisplayskip}{6pt}
	\setlength{\belowdisplayskip}{6pt}
	\small
	\bm{\tilde{h}}{^{\textit{glo}}_{i}} = \bm{h}{^{\textit{glo}}_{i}} + \sum_{j}^{m} {a_{i,j}} \bm{h}{^{\textit{glo}}_{j}},
	\end{equation}
	
	We can obtain a new local embedding $\bm{\tilde{h}}{_{i}^{\textit{loc}}}$ in a similar way.
	Consequently, we obtain refined representations for the global and local representation of sequence, denoted as
	$\bm{\tilde{h}}^{\textit{glo}} = [\bm{\tilde{h}}{_{1}^{\textit{glo}}}, ..., \bm{\tilde{h}}{_{m}^{\textit{glo}}}] $ and $\bm{\tilde{h}}^{\textit{loc}} = [\bm{\tilde{h}}{_{1}^{\textit{loc}}}, ..., \bm{\tilde{h}}{_{m}^{\textit{loc}}}] $, respectively.
	
	We  combine $\bm{\tilde{h}}^{\textit{glo}}$ and $\bm{\tilde{h}}^{\textit{loc}}$ to form a sentence representation $\bm{h}$ = $\bm{\tilde{h}}^{\textit{glo}}$ $\oplus$ $\bm{\tilde{h}}^{\textit{loc}}$.
	After that, we apply the max pooling operation to $\bm{h}$, and then concatenate it with entity representations $\bm{e^{h}}_1; ..., \bm{e^h}_n$, and finally feed them into a feed-forward neural network to form the final joint representation $\bm{h'}$, denoted as $\bm{z}$ in Figure~\ref{fig:model}:
	\begin{equation}
	\setlength{\abovedisplayskip}{6pt}
	\setlength{\belowdisplayskip}{6pt}
	\small
	\bm{z} = \bm{h'} = \textit{Linear}([f(\bm{h}); \bm{e^{h}}_1; ..., \bm{e^{h}}_n],
	\label{equ:z}
	\end{equation}
	where $f$:  $\mathbb{R}^{d \times m}$ $\to$  $\mathbb{R}^{d}$ is a max pooling function that maps \emph{m} word vectors to one vector, \emph{d} is the dimension of the vector $\bm{h}$. $\bm{e^{h}}_i$ indicates the hidden vector corresponding to the $i$-th entity in $\bm{h}$.
	
	Through the multi-scale feature learning module, we obtain two multi-scale local representation $\bm{x}$, $\bm{y}$, and one multi-scale joint relation representation $\bm{z}$. These  representations will be used in other two modules \textit{MSML} and \textit{MT}.

	\subsection{Multi-Scale Metric Learning (MSML) Module}
	The goal of metric learning is to learn a distance metric that assigns smaller   distances between similar pairs  and larger distances between dissimilar pairs. Metric learning enables the learned metric propagate the known similarity information to the pairs whose relationships are unknown, which is an essential property and plays a crucial role in   many machine learning applications.
	
	In light of this, we present the multi-scale metric learning (MSML) module to assist the RE task, including the feature-level metric learning which is used to reduce/increse the distance between the syntactic and lexical features of the same/different sample(s),  and the relation-level metric learning which aims to reduce/increse the distance between the samples in the same/different relation class(es).
	
\begin{figure}[htb]
		\center{\includegraphics[width=0.45\textwidth]{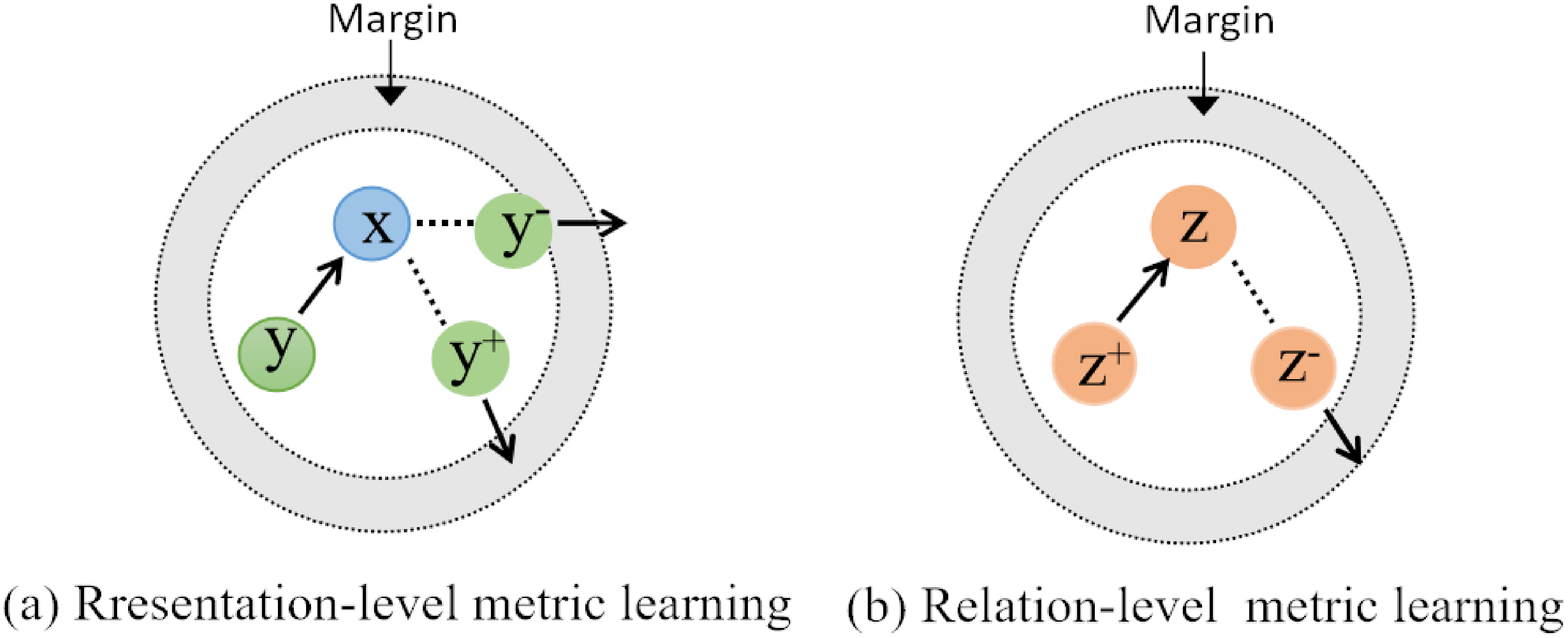}}
		\caption{An illustration example of our MSML, where `+' and `-' denote the positive and negative sample, and an `in' and `out' edge denote the pull and push operation toward the node.}
		\label{fig:msml}
		\vspace{-2mm}
	\end{figure}

	The input of the MSML module is the output anchor triplet from the MSFL module, including ($\bm{x}$, $\bm{y}$, $\bm{z}$), ($\bm{x}^+$, $\bm{y}^+$, $\bm{z}^+$), and ($\bm{x}^-$, $\bm{y}^-$, $\bm{z}^-$), where $\bm{x}$ contains the local multi-scale lexical information, $\bm{y}$ has the local multi-scale syntactic information, and $\bm{z}$ is the joint relation representation mainly used for relation classification. `$+$' and `$-$' denote  the output of positive and negative sample $S^+$ and `$S^-$', respectively.  The triplet ($\bm{z}$, $\bm{z^+}$, $\bm{z^-}$) is used for relation-level metric learning while ($\bm{x}$, $\bm{y}$, $\bm{y^+}$) and ($\bm{x}$, $\bm{y}$, $\bm{y^-}$) are used for feature-level metric learning. Figure~\ref{fig:msml} shows an  illustration example of our MSML.
	
	When selecting the negative samples, we treat the sample which is most difficult to be distinguished as the negative sample. In particular, we take the anchor $S$ as input and employ the MT module to classify the anchor samples in the validation set. From the probability distribution over relations, the relation with the second highest probability is selected as the negative sample of $S$.
	
	\subsubsection{Relation-level metric learning}
	We first apply the  relation-level metric learning to the triplet [$\bm{z}$, $\bm{z^+}$,$\bm{z^-}$] to  train our model such that   the distance between two samples in the same relation class can be reduced and that from different ones is increased. We use the triple loss as the relation loss function:
	\begin{equation}
	\setlength{\abovedisplayskip}{6pt}
	\setlength{\belowdisplayskip}{6pt}
	\small
	\textit{loss}^{\textit{rel}} = \text{max} (d(\bm{z}, \bm{z^+})^2 - d(\bm{z},\bm{z^-})^2 + m_1, 0)
	\label{equ:lossz}
	\end{equation}
	where $d(\cdot)$ is the Euclidean distance function to measure the similarity between two relation embeddings. $\text{max}(\cdot,0)$ is the hinge loss and $m_1 > 0$ is the safety margin value of the relation loss function.
	
	\subsubsection{Feature-level metric learning}
	In intuition, the local lexical representation $\bm{x}$ should be close to the local syntactic representation  $\bm{\tilde{y}}$ as they denote the same sentence. However, it is impropriate to directly let $\bm{x}$ approximate $\bm{y}$ since they contain different information. We thus perform feature-level metric learning based on the relative distance. The rationale is that the distance between local pair of the same sample should be smaller than that of different samples. This is done by applying metric learning to the representation triplet ($\bm{x}$, $\bm{y}$, $\bm{y^+}$) and ($\bm{x}$, $\bm{y}$, $\bm{y^-}$). Formally, we define the representation loss as:
	
	\begin{equation}
	\setlength{\abovedisplayskip}{6pt}
	\setlength{\belowdisplayskip}{6pt}
	\small
	\textit{loss}^{\textit{rep}}_1 = \text{max} (d(\bm{x}, \bm{y})^2 - d(\bm{x},\bm{y^+})^2 + m_2, 0)
	\label{equ:lossxyp}
	\vspace{-2mm}
	\end{equation}
	\begin{equation}
	\setlength{\abovedisplayskip}{6pt}
	\setlength{\belowdisplayskip}{6pt}
	\small
	\textit{loss}^{\textit{rep}}_2 = \text{max} (d(\bm{x}, \bm{y})^2 - d(\bm{x},\bm{y^-})^2 + m_2, 0)
	\label{equ:lossxyn}
	\vspace{-2mm}
	\end{equation}
	\begin{equation}
	\setlength{\abovedisplayskip}{6pt}
	\setlength{\belowdisplayskip}{6pt}
	\small
	\textit{loss}^{\textit{rep}} = \textit{loss}^{\textit{rep}}_1 + \textit{loss}^{\textit{rep}}_2
	\label{equ:xy}
	\end{equation}
	where $m_2 > 0$ is the safety margin value of representation loss function.

	\subsection{Main Classification (MT) Module}
	The main task in RE is to determine whether a relation holds for the entities. This task is performed on the joint representation $\bm{h'}=\bm{z}$ from the MSFL module.
	The  representation $\bm{h'}$ is sent into a softmax layer to generate the probability distribution over relations, and the loss for the main relation classification task is defined as follows:
	\begin{equation}
	\setlength{\abovedisplayskip}{6pt}
	\setlength{\belowdisplayskip}{6pt}
	\small
	\bm{p}^{h} = \textit{softmax}(\textit{Linear}(\bm{h'})) ,~
	\textit{loss}^{\textit{main}} = -\bm{p}^{h}\textit{log}\bm{\hat{p}}^{h},
	\label{main_loss}
	\end{equation}
	where $\bm{p}^{h}$ and $\bm{\hat{p}}^{h}$ are the ground truth and the predicted label distribution, respectively.

	\subsection{Model Training}
	Our model is trained with the standard gradient descent algorithm by minimizing the overall loss on all training samples, which consists of the loss for main classification and that for metric learning. Formally, the overall loss  is calculated as:
	\begin{equation}
	\setlength{\abovedisplayskip}{6pt}
	\setlength{\belowdisplayskip}{6pt}
	\small
	\textit{loss} = \sum_{i}^{J} \textit{loss}{^{\textit{main}}_{i}} + \alpha \textit{loss}{^{\textit{rel}}_{i}} + \beta \textit{loss}{^{rep}_{i}} + \lambda \Vert \Theta \Vert,
	\label{loss_all}
	\end{equation}
	where $J$ is the number of training samples, $\alpha$ and $\beta$ are hyper-parameters for controling the weights of \textit{loss}$^{\textit{rel}}$ and \textit{loss}$^{\textit{rep}}$, respectively. $\Theta$ represents all trainable parameters, and $\lambda$ is the co-efficient of L2-regularization. In the training state, we take the triplet ($S$, $S^+$, $S^-$) as input and in the test state we only use the current sample $S$ as input.

	\section{Experiments}
	\subsection{Datasets and Settings}
	\subsubsection{Datasets} We evaluate our model on two RE tasks including cross-sentence $n$-ary RE and sentence-level RE on three public datasets.
	
	For cross-sentence $n$-ary relation extraction, we use the PubMed\footnote{\href{https://github.com/freesunshine0316/nary-grn}{https://github.com/freesunshine0316/nary-grn}.} dataset ~\cite{PengPQTY17} which is from the biomedical domain. The dataset contains 6987 ternary instances about drug-gene-mutation relations, and 6087 binary instances about drug-mutation sub-relations. Most instances contain multiple sentences and each instance is assigned with one of the following five labels:  ``resistance or non-response'', ``sensitivity'', ``response'', ``resistance'' , and ``none''. Following  \cite{PengPQTY17,SongZWG18}, we consider two sub-tasks for evaluation, i.e., multi-class $n$-ary  RE and  binary-class $n$-ary RE for which we binarize multi-class labels by grouping all relation classes as ``Yes'' and treat ``None'' as ``No''.

	For sentence-level relation extraction task, we follow the experiment settings in \cite{Zhang0M18,GuoZL19} to evaluate our model on two datasets. (1) SemEval 2010 Task 8 dataset\footnote{\href{ https://github.com/Cartus/AGGCN/tree/master/semeval/dataset}{ https://github.com/Cartus/AGGCN/tree/master/semeval/dataset}.}~\cite{HendrickxKKNSPP10} has 8,000 instances for training and 2,171 instances for testing. It contains 9 directed relations and a special \textit{Other} class. (2) TACRED dataset\footnote{\href{https://nlp.stanford.edu/projects/tacred/}{https://nlp.stanford.edu/projects/tacred/}}~\cite{ZhangZCAM17} has over 106k instances and contains 41 relation types and a special \textit{no\_relation} class.
	
	Details of train/validation/test splits for PubMed, TACRED, and Semeval are shown in Table~\ref{DB}.
	\begin{table}[!h]	
		\small	
		\setlength{\abovecaptionskip}{2pt}
		\setlength{\belowcaptionskip}{0pt}
		\footnotesize
		\renewcommand\arraystretch{1.1}
		\centering
		\caption{Train/validation/test splits for PubMed, TACRED and SemEval. ``C'' and ``I'' denotes the number of categories and instances, respectively.}
		\vspace{-3mm}
		\begin{tabular}{lcccccc} \\
			\toprule[1pt]
			\multirow{2}[0]{*}{\textbf{Dataset}} & \multicolumn{2}{c}{\textbf{PubMed}} & \multicolumn{2}{c}{\textbf{TACRED}} & \multicolumn{2}{c}{\textbf{Semeval}}\\
			\cmidrule(r){2-3} \cmidrule(r){4-5}	\cmidrule(r){6-7}
			& C & I & C & I &	C & I\\
			\hline
			Train&	5&	5313 &42 &68124 &10 &8000\\
			\hline
			Validation& 5&200 &42 & 22631&- &-\\
			\hline
			Test&	5&	1474 &42 & 15509&10 &2717\\
			\midrule[1pt]		
		\end{tabular}%
		\vspace{-2mm}
		\label{DB}%
	\end{table}%

	\subsubsection{Settings}
	For  models with static word embedding, we initialize word vectors with the 300-dimension GloVe embeddings provided by \cite{PenningtonSM14}. The embeddings for POS and position are initialized randomly and their dimension is set to 30~\cite{GuoZL19}. We use the Stanford CoreNLP~\cite{ManningSBFBM14} as the parser and SGD as the optimizer with a 0.9 decay rate. The L2-regularization coefficient $\lambda$ is 0.002.
	
	For models with contextualized  word embedding, we use $\text{BERT}_{\text{base}}$~\cite{DevlinCLT19}, $\text{BERT}_{\text{large}}$~\cite{DevlinCLT19}, $\text{RoBERTa}$~\cite{Liu2020Myle}, and  $\text{RoBERTa}_{\text{large}}$~\cite{Liu2020Myle} as  pre-trained language models (PLMs), respectively.
	
	For cross-sentence $n$-ary RE, we use the same data split as that in~\cite{SongZWG18}. Through preliminary experiments on the development set, we adopt the following  combinations of the hyper-parameters ($\alpha$, $\beta$) as they yield the best results: $\alpha$ = 0.7 and $\beta$ = 0.03 for ternary-relation multi-class, $\alpha$ = 0.5, and $\beta$ = 0.03 for ternary-relation binary-class, $\alpha$ = 0.1 and $\beta$ = 0.07 for binary-relation multi-class and $\alpha$ = 0.1 and $\beta$ = 0.09 for binary-relation binary-class.
	
	For sentence-level RE, following \cite{Zhang0M18,GuoZL19}, we use the validation set in TACRED and do not use the validation set in SemEval but test on the last epoch's checkpoint among 150 training epochs. The best results are based on the combinations of hyper-parameters, i.e., $\alpha$ = 0.7 and $\beta$ = 0.05 for TACRED and $\alpha$ = 0.5 and $\beta$ = 0.03 for Semeval.
	
	\subsubsection{Evaluation metrics}
	We use the same metrics as previous work~\cite{SongZWG18,Zhang0M18,GuoZL19}  to evaluate our model. In particular, the test accuracy averaged over five-fold cross validation ~\cite{SongZWG18} is used  as the metric for the cross-sentence $n$-ary RE on PubMed, and micro- and macro- averaged F1 scores are used for sentence level RE on TACRED and SemEval dataset, respectively.

	\begin{table*}[h]
		\vspace{-0mm}
		\small
		\centering	
		\caption{Comparison results in terms of accuracy on PubMed. ``T'' and ``B'' denote the ternary and binary entity interactions, and ``Single'' and ``Cross'' mean the accuracy calculated within single sentences or on all sentences. The best results with GloVe/BERT embeddings are both in bold, and the second best ones with GloVe/BERT embeddings are underlined/italic. $\dagger$  and $\ddagger$  mark denote statistically significant improvements over the corresponding second best results with $p <$ .05 and $p <$ .01, respectively.}
		\vspace{-4mm}
		\begin{tabular}{lllllll} \\
			\toprule[1pt]
			\multirow{3}[0]{*}{\textbf{Model}} & \multicolumn{4}{c}{\textbf{Binary-class}} & \multicolumn{2}{c}{\textbf{Multi-class}}  \\
			\cmidrule(r){2-5} \cmidrule(r){6-7}
			&\multicolumn{2}{c}{\textbf{T}} & \multicolumn{2}{c}{\textbf{B}} & \textbf{T} & \textbf{B}\\	
			& Single  & Cross  & Single  & Cross   & Crss  & Cross\\
			\hline		PA-LSTM~\cite{ZhangZCAM17}    & 84.9 & 85.8  & 85.6 & 85.0 & 78.1 & 77.0 \\
			\hline
			Feature-Based~\cite{QuirkP17} & 74.7 & 77.7 & 73.9 & 75.2 & - & -  \\	
			\hline	
			SP-Tree~\cite{MiwaB16} &- &-	&75.9	&75.9	&-	&-\\
			Graph LSTM-EMBED~\cite{PengPQTY17}   & 76.5 & 80.6 & 74.3 & 76.5 & - & - \\
			Graph LSTM-FULL~\cite{PengPQTY17}   & 77.9 & 80.7 & 75.6 & 76.7 & - & - \\
			Bidir DAG LSTM~\cite{SongZWG18}	& 75.6	& 77.3	& 76.9	& 76.4	& 51.7	& 50.7 \\
			GS GLSTM~\cite{SongZWG18}	& 80.3	& 83.2&	83.5&	83.6&	71.1&	71.7\\
			\hline	
			GCN (Full Tree)~\cite{Zhang0M18} &84.3	&84.8	&84.2	&83.6	&77.5	&74.3 \\
			GCN ($K$ = 0)~\cite{Zhang0M18} &85.8	&85.8	&82.8	&82.7	&75.6	&72.3 \\
			GCN ($K$ = 1)~\cite{Zhang0M18} &85.4 	&85.7	&83.5 	&83.4	&78.1	&73.6 \\
			GCN ($K$ = 2)~\cite{Zhang0M18} &84.7	&85.0	&83.8	&83.7	&77.9	&73.1 \\
			AGGCN~\cite{GuoZL19}         &87.1	&87.0 	&85.2	&85.6	&79.7 	&77.4 \\
			EDCRE~\cite{LeeSOLSL20} &87.9	&\underline{89.0}	&86.4	&86.8	&\underline{82.5}	&\underline{79.9} \\
			LF-GCN~\cite{GuoN0C20} &\underline{88.0}	&88.4	&\underline{86.7}	&\underline{87.1}	&81.5	&79.3 \\
			C-GCN-MG~\cite{MandyaBC20}   &87.2	&88.5	&86.1	&86.9	&82.1	&78.8 \\
			C-GCN-MG (BERT$_\text{base}$)~\cite{MandyaBC20} &{\textit{88.3}}	&{\textit{88.1}}	&{\textit{87.9}}	&{\textit{87.2}}	&{\textit{86.2}}	&{\textit{87.3}} \\			
			\hline		
			Our model (GloVe)&\textbf{89.1}$^{\ddagger}$	&\textbf{89.3}$^{\dagger}$	&\textbf{87.7}$^{\ddagger}$	&\textbf{87.5}$^{\ddagger}$	&\textbf{84.6}$^{\ddagger}$	&\textbf{81.6}$^{\ddagger}$\\		
			Our model (BERT$_\text{base}$)
			&\textbf{90.2}$^{\ddagger}$
			&\textbf{90.7}$^{\ddagger}$	&\textbf{90.5}$^{\ddagger}$	&\textbf{89.5}$^{\ddagger}$	&\textbf{87.9}$^{\ddagger}$	&\textbf{89.1}$^{\ddagger}$ \\
			Our model (BERT$_\text{large}$) 		&\textbf{91.1}$^{\ddagger}$         &	\textbf{91.3}$^{\ddagger}$&	\textbf{91.6}$^{\ddagger}$&	\textbf{91.5}$^{\ddagger}$&	\textbf{88.6}$^{\ddagger}$&	\textbf{89.6}$^{\ddagger}$\\
			Our model (RoBERTa$_\text{base}$) 		&\textbf{91.8}$^{\ddagger}$&	\textbf{91.9}$^{\ddagger}$&	\textbf{92.4}$^{\ddagger}$&	\textbf{92.3}$^{\ddagger}$&	\textbf{89.5}$^{\ddagger}$&	\textbf{90.4}$^{\ddagger}$ \\
			Our model (RoBERTa$_\text{large}$) 		&\textbf{92.6}$^{\ddagger}$&	\textbf{92.4}$^{\ddagger}$&	\textbf{92.8}$^{\ddagger}$&	\textbf{92.9}$^{\ddagger}$&	\textbf{90.3}$^{\ddagger}$&	\textbf{90.8}$^{\ddagger}$\\
			\midrule[1pt]		
		\end{tabular}%
		\label{tab1}%
	\end{table*}%
	
	\subsection{Results for Cross-Sentence $n$-ary RE}
	\subsubsection{Baselines}
	For cross-sentence $n$-ary relation extraction task, we compare our model with following baselines which are grouped into two categories.
	
	\textbf{Sequence based models}: PA-LSTM ~\cite{ZhangZCAM17} is a classic model applying LSTM to the word sequence where each word embedding contains word,  position, and POS tag.
	
	\textbf{Dependency based models}: We compare our model with three types of dependency based modes.
	(1) Feature-based ~\cite{QuirkP17} is a statistical method with features derived from the shortest paths in dependency tree.
	(2) Tree-based methods: SP-Tree ~\cite{MiwaB16} is a tree-structured LSTM method. Graph LSTM-EMBED ~\cite{PengPQTY17}, Graph LSTM-FULL ~\cite{PengPQTY17}, Bidir DAG LSTM ~\cite{SongZWG18}, and GS GLSTM ~\cite{SongZWG18} are LSTM models that adopt graph-structure with dependency as edges.
	(3) Graph-based methods: GCN~\cite{Zhang0M18} pools information over pruned trees. AGGCN~\cite{GuoZL19}  takes full dependency trees as inputs with a soft-pruning approach. EDCRE~\cite{LeeSOLSL20} considers both dependency edges between entities and discourse relations between sentences. LF-GCN~\cite{GuoN0C20} composes task-specific dependency forests that capture non-local interactions.
	C-GCN-MG~\cite{MandyaBC20} performs GCN operation over multiple dependency sub-graphs and it is the only baseline using BERT to get initial word embeddings.
	
	We re-implement PA-LSTM, EDCRE, and C-GCN-MG (with GloVe embedding) using the optimal hyper-parameter settings reported in corresponding papers since they do not provide full results for all tasks in our experiments. Results for other baselines are taken from the original papers.
	
	\subsubsection{Main results}
	The comparison results for all methods are shown in Table~\ref{tab1}. From these results, we make the following observations.
	
	(1) Our model achieves the best accuracy in cross-sentence $n$-ary RE tasks with both GloVe and BERT embeddings. In particular, it gets an improvement of 2.1\% and 1.7\% accuracy over the second best one on ternary and binary relation multi-class RE, respectively. Moreover, with the $\text{BERT}_{\text{base}}$ embedding, our model significantly outperforms its counterpart baseline C-GCN-MG~\cite{MandyaBC20} with the same embedding.
	
	(2) For baselines with contextualized word embeddings, their results significantly outperform the methods with static word embeddings, which is reasonable. In addition, their performances increase with the size of corpus ($\text{BERT}_{\text{large}}$ $>$ $\text{BERT}_{\text{base}}$, $\text{RoBERTa}_{\text{large}}$ $>$ $\text{RoBERTa}_{\text{base}}$), and they also increase with the performance of PLMs ($\text{RoBERTa}$ $>$ $\text{BERT}$). Both these prove the power of PLMs beyond the methods themselves. They also indicate that a fair comparison between PLM based methods should be conducted using the same PLM, e.g.,  $\text{BERT}_{\text{base}}$ $vs.$ $\text{BERT}_{\text{base}}$.
	
	(3) For baselines with static word embeddings, EDCRE and LF-GCN using graphs are better than those using LSTM among dependency based models,  indicating that the graph structure is more appropriate for modeling dependency trees than the sequential structure. Moreover, the better performance of LF-GCN can be due to the fact that it develops a new method to obtain task-specific dependency relations. The sequence based model PA-LSTM  outperforms the dependency based methods using LSTM and even matches those with GCN, indicating that both word sequence and parse tree have their own strength given the proper architecture to extract lexical and syntactic patterns.
	

	\subsection{Results for Sentence-level RE}
	\subsubsection{Baselines}
	We adopt the popular baselines for sentence-level RE task and group them into three categories.
	
	\textbf{Sequence based models}: We consider the state-of-art PA-LSTM ~\cite{ZhangZCAM17} as the representative of sequence based models.
	
	\textbf{Dependency based models}:
	(1) tree-based methods:
	the shortest path LSTM (SDP-LSTM) ~\cite{XuMLCPJ15},
	the tree-structure neural model Tree-LSTM ~\cite{TaiSM15} and the classic LR classifier over dependency features ~\cite{ZhangZCAM17} for TACRED dataset,
	SVM ~\cite{RinkH10} model combines various linguistic resources for recognizing semantic relations in text.
	SP-Tree ~\cite{MiwaB16} model employs bidirectional tree-structured LSTM-RNNs to capture syntactic information on Semeval dataset.
	(2) graph-based methods: C-GCN ~\cite{GuoZL19} with pruned trees, C-AGGCN ~\cite{GuoZL19} extends the original methods by using contextualized embeddings with a Bi-LSTM as input. EDCRE~\cite{LeeSOLSL20}, LF-GCN~\cite{GuoN0C20}, and C-GCN-MG~\cite{MandyaBC20} are same as those in cross-sentence $n$-ary RE.
	
	\textbf{PLM based models}:
	MTB~\cite{SoaresFLK19} is post-trained by classifying whether two sentences mention the same entity pair with entity mentions randomly masked which uses Wikidata for pre-training. CP~\cite{PengGHLLLSZ20} is an entity-masked contrastive post-training framework for RE which also uses Wikidata for pre-training. MTB~\cite{SoaresFLK19} and  CP~\cite{PengGHLLLSZ20} both report the results on TACRED and Semeval.
	LUKE~\cite{YamadaASTM20} is post-trained to  predict randomly masked words and entities in a large entity-annotated corpus retrieved from Wikipedia which employs RoBERTa$_\text{large}$ as transformers for TACRED.
	R-BERT~\cite{WuH19a} with BERT$_\text{base}$ and BERT$_\text{large}$  incorporates information from the target entities to tackle the relation classification task on Semeval via fine-tuning.
	Note that, we do not use extra corpus to help relation extraction task in our model. Our proposed  post-trained tasks and relation extraction task are both conducted on the same RE dataset. The results for PLM based baselines are taken from the original papers.
	
	\subsubsection{Main results}
	We present the main results on TACRED in Table~\ref{tab2} and Semeval in Table~\ref{tab3}.
	
	Our model achieves the best results in terms of Micro-F1 on TACRED or Macro-F1 scores on Semeval with GloVe embeddings. Other results are also consistent with those in cross sentence $n-$ary task.
	
	For PLM based models, we can see that our model beats all the baseline methods, no matter the input contextualized word embedding  is from BERT$_\text{base}$, BERT$_\text{large}$, RoBERTa$_\text{base}$, or RoBERTa$_\text{large}$.
	Note that our model with contextualized embeddings gets less improvement over the second best baselines (comparing F1 score BERT$_\text{large}$ 71.7 \emph{vs}. 71.5, RoBERTa$_\text{large}$ 73.1 \emph{vs}. 72.7 in Table~\ref{tab2}). The reason might be that we do not use the extra Wikidata for post-training the word embeddings. Nevertheless, our model can already outperform the baselines  with  the extra large corpus. This clearly shows the effectiveness of our model.
	

	
	\begin{table}[h]	
		\small	
		\setlength{\abovecaptionskip}{2pt}
		\setlength{\belowcaptionskip}{0pt}
		\footnotesize
		\renewcommand\arraystretch{1.1}
		\centering
		\caption{Micro-avg. precision, recall, and F1 on TACRED. The marks are as same as those in Table ~\ref{tab1}. Note that there is no significance test for PLM based models since the results for baselines are taken from the original papers.}
		\begin{tabular}{llll} \\
			\toprule[1pt]
			\textbf{Model} & \textbf{P}& \textbf{R}& \textbf{F1} \\
			\hline
			PA-LSTM~\cite{ZhangZCAM17} &	65.7&	64.5&	65.1 \\
			\hline
			Tree-LSTM~\cite{TaiSM15}&	66.0&	59.2&	62.4\\
			LR~\cite{ZhangZCAM17} &73.5 &	49.9 &	59.4 \\		
			SDP-LSTM~\cite{XuMLCPJ15} &	66.3&	52.7&	58.7 \\
			\hline
			C-GCN~\cite{Zhang0M18} &69.9&	63.3&	66.4\\
			C-AGGCN~\cite{GuoZL19}	& 71.9 &64.0 &	\underline{67.7} \\
			EDCRE~\cite{LeeSOLSL20} &71.2	&63.3 &	67.0 \\	
			C-GCN-MG ~\cite{MandyaBC20} &	67.5 &	63.1 &	65.2 \\
			\hline
			MTB (BERT$_\text{base}$) ~\cite{SoaresFLK19} & - & - & 69.1 \\
			MTB (BERT$_\text{large}$) ~\cite{SoaresFLK19} & - & - & \textit{71.5} \\
			C-GCN-MG (BERT$_\text{base}$) \cite{MandyaBC20} &68.0 &	64.4 &66.1  \\
			CP (BERT$_\text{base}$)~\cite{PengGHLLLSZ20} & - & - & \textit{69.5} \\
			LUKE (RoBERTa$_\text{large}$)~\cite{YamadaASTM20} & 70.4 & 75.1 &\textit{72.7}\\
			\hline		
			Our model (GloVe) &72.9	&64.3 & \textbf{68.3}$^{\dagger}$ \\		
			Our model (BERT$_\text{base}$) &73.6 &	67.5 &\textbf{70.5}  \\
			Our model (BERT$_\text{large}$) &73.4 &	70.1 &	\textbf{71.7}  \\
			Our model (RoBERTa$_\text{base}$) & 71.8&	73.7&	\textbf{72.7}	 \\
			Our model (RoBERTa$_\text{large}$) &72.1 &	74.2 &	\textbf{73.1} \\
			\midrule[1pt]		
		\end{tabular}%
		\label{tab2}%
	\end{table}%
	
	\begin{table}[h]
		\small	
		\setlength{\abovecaptionskip}{3pt}
		\setlength{\belowcaptionskip}{0pt}
		\footnotesize
		\renewcommand\arraystretch{1.1}
		\centering
		\caption{ Macro-avg. F1 on SemEval. The marks are as same as those in Table ~\ref{tab1}. Note that there is no significance test for PLM based models since the results for baselines are taken from the original papers.}
		\begin{tabular}{ll} \\		
			\toprule[1pt]
			\textbf{Model} & \textbf{F1} \\
			\hline
			PA-LSTM~\cite{ZhangZCAM17} &	82.7 \\
			\hline
			SVM~\cite{RinkH10}&	82.2 \\
			SDP-LSTM~\cite{XuMLCPJ15} &	83.7 \\
			SP-Tree~\cite{MiwaB16} & 84.4 \\
			\hline
			C-GCN~\cite{Zhang0M18} &	84.8 \\
			C-AGGCN~\cite{GuoZL19}	& 85.7 \\
			EDCRE~\cite{LeeSOLSL20} & \underline{85.9} \\
			LF-GCN~\cite{GuoN0C20} & 85.7 \\
			C-GCN-MG~\cite{MandyaBC20} &	82.4 \\
			\hline
			MTB (BERT$_\text{base}$) ~\cite{SoaresFLK19} & 87.3 \\
			MTB (BERT$_\text{large}$) ~\cite{SoaresFLK19} & \textit{89.5} \\
			R-BERT (BERT$_\text{base}$) ~\cite{WuH19a} & \textit{88.4} \\
			R-BERT (BERT$_\text{large}$) ~\cite{WuH19a} & 89.3 \\
			C-GCN-MG (BERT$_\text{base}$)~\cite{MandyaBC20}& 85.9\\
			CP (BERT$_\text{base}$)~\cite{PengGHLLLSZ20} & 87.6 \\
			\hline		
			Our model (GloVe)&\textbf{86.4}$^{\dagger}$  \\		
			Our model (BERT$_\text{base}$) &\textbf{89.1}  \\
			Our model (BERT$_\text{large}$) &\textbf{89.8} \\
			Our model (RoBERTa$_\text{base}$)& \textbf{90.3} \\
			Our model (RoBERTa$_\text{large}$)& \textbf{91.8} \\
			\midrule[1pt]		
		\end{tabular}
		\vspace{-3mm}
		\label{tab3}%
	\end{table}%

	\subsection{Ablation study}
	
	To examine the impacts of each component in our model, we further  conduct two types of ablation study. One is for validating the effects of three components (MSCNN, MSGCN, and MSML). This is done by removing the whole component from the model.  The other is for validating the effects of multi-scale learning. This is done by replacing MSCNN/MSGCN with the ordinary CNN/GCN.
	We present the ablation results for  cross-sentence $n$-ary RE on PubMed in Table~\ref{abl:pubmed_without_replace}, and those for  sentence-level RE on TACRED and SemEval dataset in Table~\ref{abl:tacred_without_replace}, respectively.
	\begin{table*}[h]
		\small
		\centering	
		\caption{Results for ablation study(\%) on PubMed dataset. M1-M6 are models with GloVe embeddings. M7-M12 are models with BERT$_\text{base}$ embeddings. $w/o$ denotes removing the whole component and $r/p$ denotes replacing the MSCNN/MSGCN with the ordinary CNN/GCN.}
		\vspace{-4mm}
		\begin{tabular}{lllllll} \\
			\toprule[1pt]
			\multirow{3}[0]{*}{\textbf{Model}} & \multicolumn{4}{c}{\textbf{Binary-class}} & \multicolumn{2}{c}{\textbf{Multi-class}}  \\
			\cmidrule(r){2-5} \cmidrule(r){6-7}
			&\multicolumn{2}{c}{\textbf{T}} & \multicolumn{2}{c}{\textbf{B}} & \textbf{T} & \textbf{B}\\	
			& Single  & Cross  & Single  & Cross   & Cross  & Cross\\
			\hline	
			M1$^{-}$ w/o MSCNN & 87.8&	87.8&	86.7&	86.0&	82.9&	79.1	 \\
			M2$^{-}$ w/o  MSGCN & 87.3&	87.5&	86.4&	85.8&	82.7&	79.0 	 \\	
			M3$^{-}$ w/o MSML    & 88.9& 89.1 & 87.4 & 86.6 & 84.1 & 80.9 \\
			M4$^{-}$  r/p MSCNN &88.4	&88.6&	87.4&	86.4&	83.7&	80.8 \\
			M5$^{-}$  r/p MSGCN	& 88.3	&88.5&	87.2&	86.2&	83.4&	80.6  \\		
			M6 Our  model (GloVe) &\textbf{89.1}	&\textbf{89.3}	&\textbf{87.7}	&\textbf{87.5}	&\textbf{84.6}	&\textbf{81.6}\\
			\hline
			M7$^{-}$ w/o MSCNN &89.7&	89.6&	89.6&	88.8&	87.3&	88.6 \\
			M8$^{-}$ w/o MSGCN &89.7&	89.7&	89.8&	88.6&	87.1&	88.4 \\
			M9$^{-}$ w/o MSML & 89.8	&89.6&	89.9&	89.1&	87.8&	88.8\\
			M10$^{-}$ r/p MSCNN &89.9&	90.4&	90.3&	89.2&	87.8&	88.9 \\
			M11$^{-}$ r/p MSGCN &89.8&	90.3&	90.4&	89.4&	87.6&	88.9 \\
			M12 Our  model (BERT$_\text{base}$)& \textbf{90.2}	&\textbf{90.7}&	\textbf{90.5}&	\textbf{89.5}&	\textbf{87.9}&\textbf{89.1} \\
			
			\midrule[1pt]		
		\end{tabular}%
		\label{abl:pubmed_without_replace}%
	\end{table*}%

	\begin{table}[!h]	
		\small	
		\setlength{\abovecaptionskip}{2pt}
		\setlength{\belowcaptionskip}{0pt}
		\footnotesize
		\renewcommand\arraystretch{1.1}
		\centering
		\caption{Results for ablation study(\%) on TACRED and SemEval dataset. M1-M6 are models with GloVe embeddings. M7-M12 are models with BERT$_\text{base}$ embeddings. $w/o$ denotes removing the whole component and $r/p$ denotes replacing the MSCNN/MSGCN with the ordinary CNN/GCN.}
		\begin{tabular}{lllcc} \\
			\toprule[1pt]
			\multirow{2}[0]{*}{\textbf{Model}} & \multicolumn{3}{c}{\textbf{TACRED}}&
			\textbf{SemEval}\\
			& \textbf{P}& \textbf{R}& \textbf{Micro-F1} & \textbf{Macro-F1}\\
			\hline
			M1$^-$ w/o MSCNN 	&72.3&	63.1&	67.4& 85.5\\
			M2$^-$  w/o MSGCN  &72.1&	62.8&	67.1 & 85.3\\
			M3$^-$ w/o MSML  &72.5&	63.7&	67.8& 86.2\\
			M4$^-$ r/p MSCNN &72.4&	63.4&	67.6 &86.1\\
			M5$^-$  r/p MSGCN  &72.5&	63.2&	67.5& 85.9\\
			M6 Our  model (GloVe) &72.9	&64.3&	\textbf{68.3} & \textbf{86.4}\\	
			\hline
			M7$^-$  w/o MSCNN &73.4	&67.2&	69.7 & 88.6 \\
			M8$^-$  w/o MSGCN &73.1	&67.3&	69.6 & 88.5 \\		
			M9$^-$  w/o MSML &73.3&	67.2&	70.3 & 88.8\\
			M10$^-$  r/p MSCNN &73.4 &67.1&	70.1 & 88.9\\
			M11$^-$  r/p MSGCN  &73.2&	67.0&	70.0 & 88.8\\		
			M12 Our  model (BERT$_\text{base}$) & 73.6&	67.5&\textbf{70.5}&\textbf{89.1} \\	
			\midrule[1pt]		
		\end{tabular}%
		\label{abl:tacred_without_replace}%
	\end{table}%

	From the results in Table~\ref{abl:pubmed_without_replace} and Table~\ref{abl:tacred_without_replace}, we find both the removal and replacement operations incur the performance  decrease in comparison with their counterparts, which clearly demonstrates that our three components and the multi-scale learning strategy contribute to our model.
	
	Among three components, the removal of MSGCN results in the biggest performance decrease. For example,  with GloVe embeddings, the complete model M6 has an accuracy score 81.6 on PubMed (Multi-class, B) in Table~\ref{abl:pubmed_without_replace}. After removing MSGCN, the accuracy for the reduced method M2$^-$ drops to 79.0, showing a 1.6 absolute decrease. Similarly, the replacement of MSGCN on the same task also brings about an 1.0 accuracy decrease (from M6 81.6 to M5$^-$ 80.6). This clearly proves that our proposed MSGCN enlarges  the words' receptive field in the parse tree and shows impressive performance enhancement.
	
	Between the methods with two types of word embeddings, our components have larger positive impacts on the static GloVe embeddings than the contextualized BERT embeddings. For example, on SemEval dataset, the removal of MSCNN for GloVe brings about a 0.9 F1 decrease (from M6 86.4 to M1$^-$ 85.5) in Table~\ref{abl:tacred_without_replace}. In contrast, the removal of MSCNN for BERT$_\text{base}$ only has a  0.5 F1 decrease (from M12 89.1 to M1$^-$ 88.6). Similar results can be observed for MSGCN and MSML components. The reason might be that BERT has already learned sufficient information from its own multiple transformer layers and the space for improvements is limited, especially on those small datasets like SemEval. Nevertheless, on the larger dataset like TACRED, our multiscale feature and metric learning paradigm still shows promising improvements.

	\subsection{Case Study}
	To have a close look, we select three examples from three datasets for a case study and present the results of different models in Table~\ref{case}. We choose PA-LSTM (the STOA sequence based model), EDCRE (the best dependency based model) with GloVe embeddings, and C-GCN-MG and R-BERT  with BERT$_\text{base}$ embeddings as the comparison methods for case study.
	
	
	The first example is from the PubMed dataset. It generally mentions that `cetuximab' (drug) does not have an effect on `S492R' (mutation) on `EGFR' (gene).
	The context introduces more ambiguity by mentioning that another drug `panitumumab' is related to `EGFR' and `S492R'. Hence both PA-LSTM and EDCRE fail to extract the correct relation between entities. In contrast, our model connects `cetuximab' with `not' via the dilated horizontal convolution operation in MSCNN and hence it makes a correct prediction.
	
	The second example  is  from the TACRED dataset. It aims at detecting the relation `org:alternate\_names' between two organizations `Lashkar-e-Taiba' (subject) and `Army of the Pure' (object). Both entities have more than two words, and the sentence is too long for a ordinary CNN and GCN in in PA-LSTM or C-GCN-MG$_B$ to capture the relation between two entities. Our model, on the other hand, enlarges the receptive field for each word with the help of MSCNN and MSGCN, and finally our model successfully recognizes the  `org:alternate\_names' relation.

	\begin{table}[!h]	
		\small	
		\setlength{\abovecaptionskip}{2pt}
		\setlength{\belowcaptionskip}{0pt}
		\footnotesize
		\renewcommand\arraystretch{1}
		\centering
		\caption{Case study. The left column presents the selected examples, and the two columns on the right denote the extraction results of corresponding models. Words in red are entities. ``G'' and ``B'' indicate models with GloVe and BERT$_\text{base}$ embeddings, respectively.
		}
		
		\begin{tabular}{l|ll}
			\hline
			\textbf{Example}  & \textbf{Model}  &
			\textbf{Prediction}
			
			\\
			\hline
			\multirow{4}{*}{\tabincell{l}{Panitumumab can still bind to an \\ \textcolor{red}{EGFR} mutant \textcolor{red}{S492R} to which\\ \textcolor{red}{cetuximab} can not bind to.}} & PA-LSTM$_\text{G}$ & response \XSolidBrush \\
			& EDCRE$_\text{G}$ & response \XSolidBrush \\
			& Our model$_\text{G}$ & none \Checkmark \\
			& Our model$_\text{B}$ & none \Checkmark \\
			\hline
			\multirow{4}{*}{\tabincell{l}{But US and Indian experts say it \\ has hesitated to take action\\ against  \textcolor{red}{Lashkar-e-Taiba} ,  which \\means ``The \textcolor{red}{Army of the Pure}, ''  \\believing that the Islamic militants\\ ... rival India.
			}} & PA-LSTM$_\text{G}$ &  no\_relation \XSolidBrush \\
			& C-GCN-MG$_\text{B}$ & member\_of \XSolidBrush \\
			& Our model$_\text{G}$ & alternate\_names \Checkmark\\
			&Our model$_\text{B}$ & alternate\_names \Checkmark \\
			\\
			\\
			\hline
			\multirow{4}{*}{\tabincell{l}{They connected the \textcolor{red}{plug} of \\ the \textcolor{red}{battery box} to the plug \\ of the charger .
			}} & PA-LSTM &  Compo-Whole \XSolidBrush \\
			& R-BERT$_\text{B}$ & Compo-Whole \XSolidBrush \\
			& Our model$_\text{G}$ & Other \Checkmark\\
			&Our model$_\text{B}$ & Other \Checkmark \\
			\hline				
		\end{tabular}%
		\label{case}%
	\end{table}%

	The third example is from the SemEval dataset. Its goal is to detect the relation  between `plug' and `battery box', which does not exist. PA-LSTM and R-BERT$_B$ treat `plug' to be a  part of `battery box' and then assign the `Component-Whole' relation. Our reduced model without MSML also makes the same wrong prediction. However, our complete model can successfully recognize the `other' relation. This clearly proves the effectiveness of our multi-scale metric learning scheme.

	\subsection{Parameter Analysis}
\begin{figure}[!htb]
		\centering
		\small
		\begin{minipage}[t]{0.49\linewidth}
			\centering
			\includegraphics[width=1\textwidth]{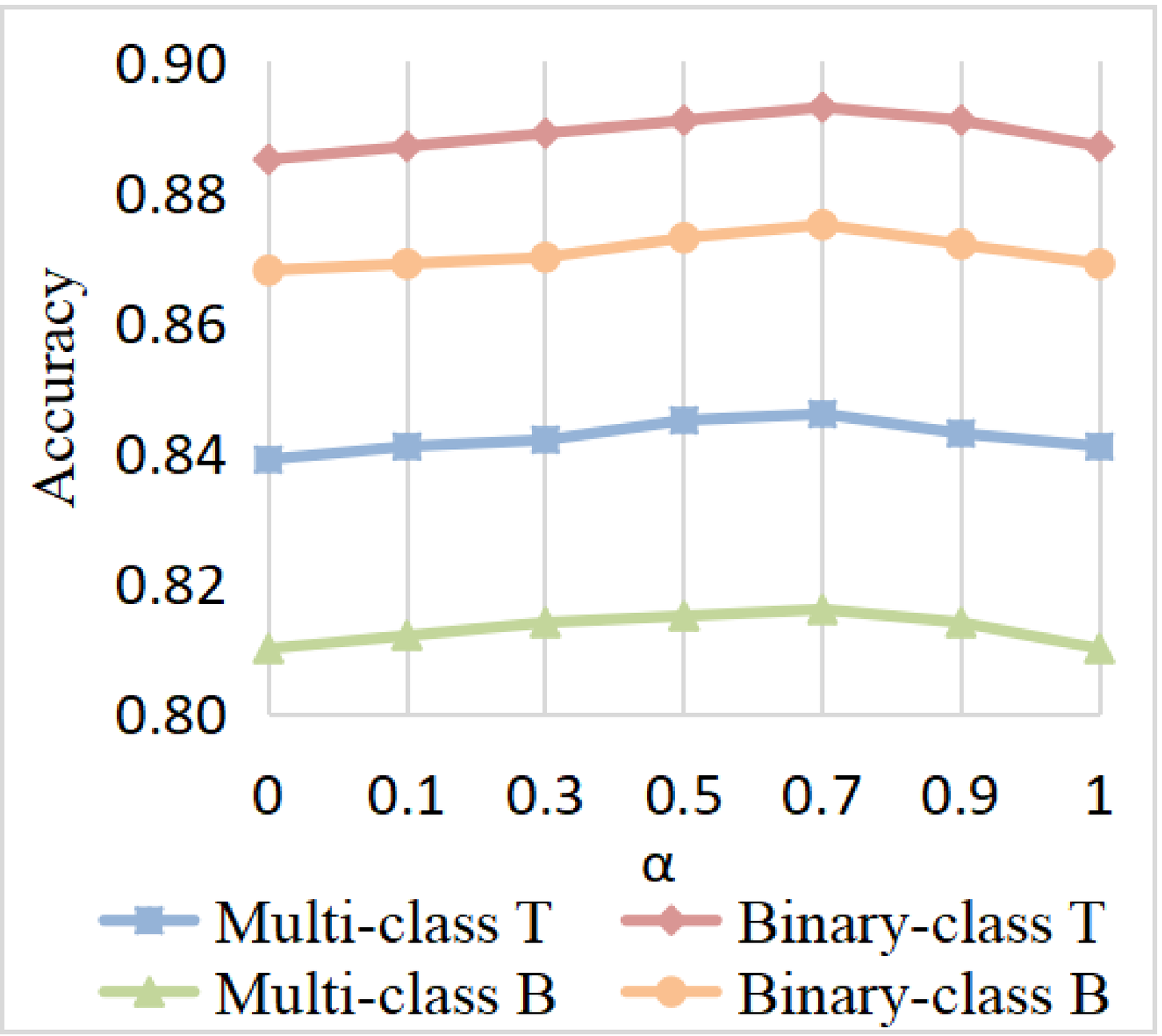}
			\subcaption{$\alpha$}
		\end{minipage}
		\begin{minipage}[t]{0.49\linewidth}
			\centering
			\includegraphics[width=1\textwidth]{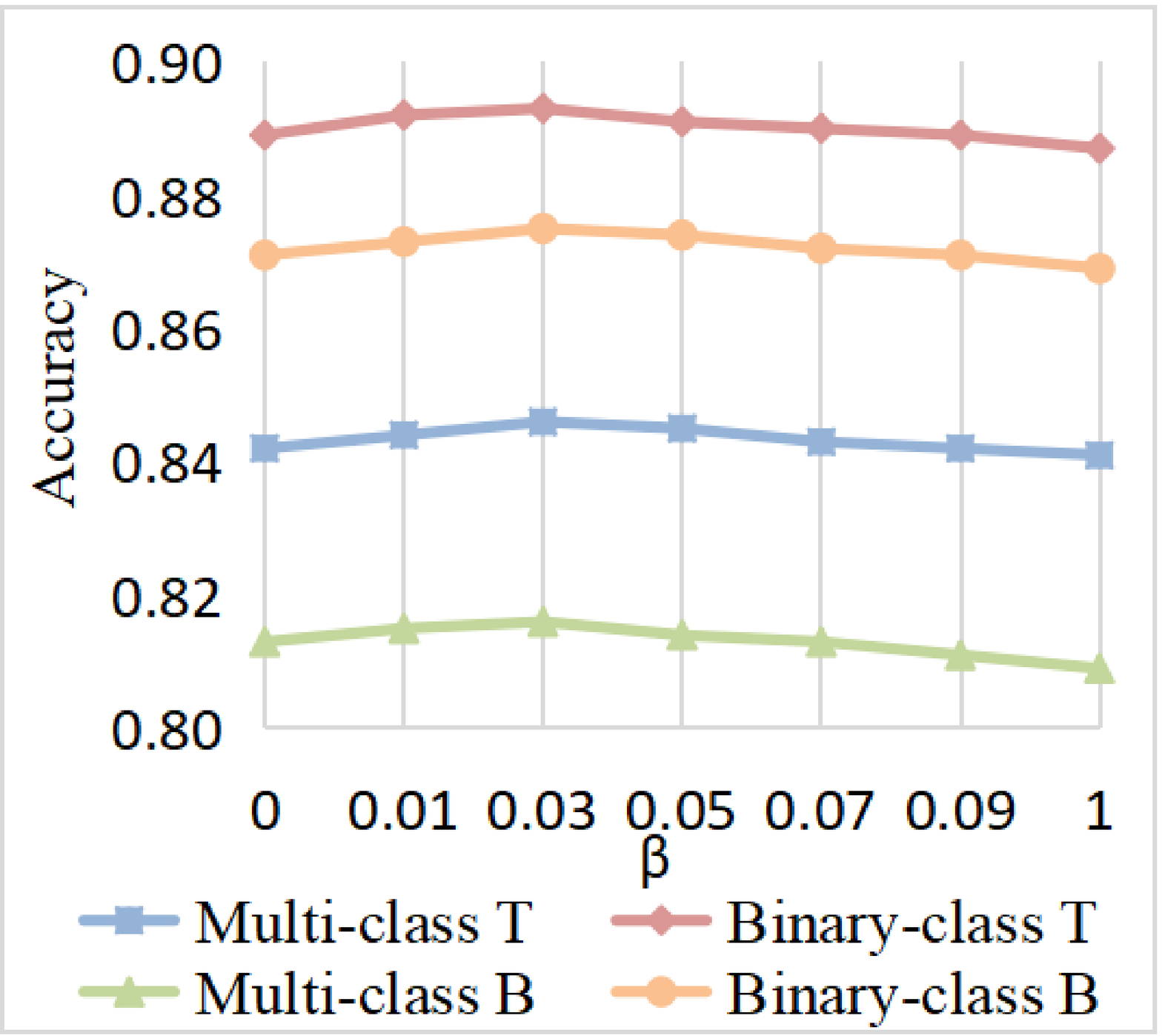}
			\subcaption{$\beta$}
		\end{minipage}
		\caption{Impacts of two hyper-parameters $\alpha$ and $\beta$ on PubMed.}	
		\label{fig:para}
	\end{figure}

\begin{figure}[!htb]
		\centering
		\small
		\begin{minipage}[t]{0.49\linewidth}
			\centering
			\includegraphics[width=1\textwidth]{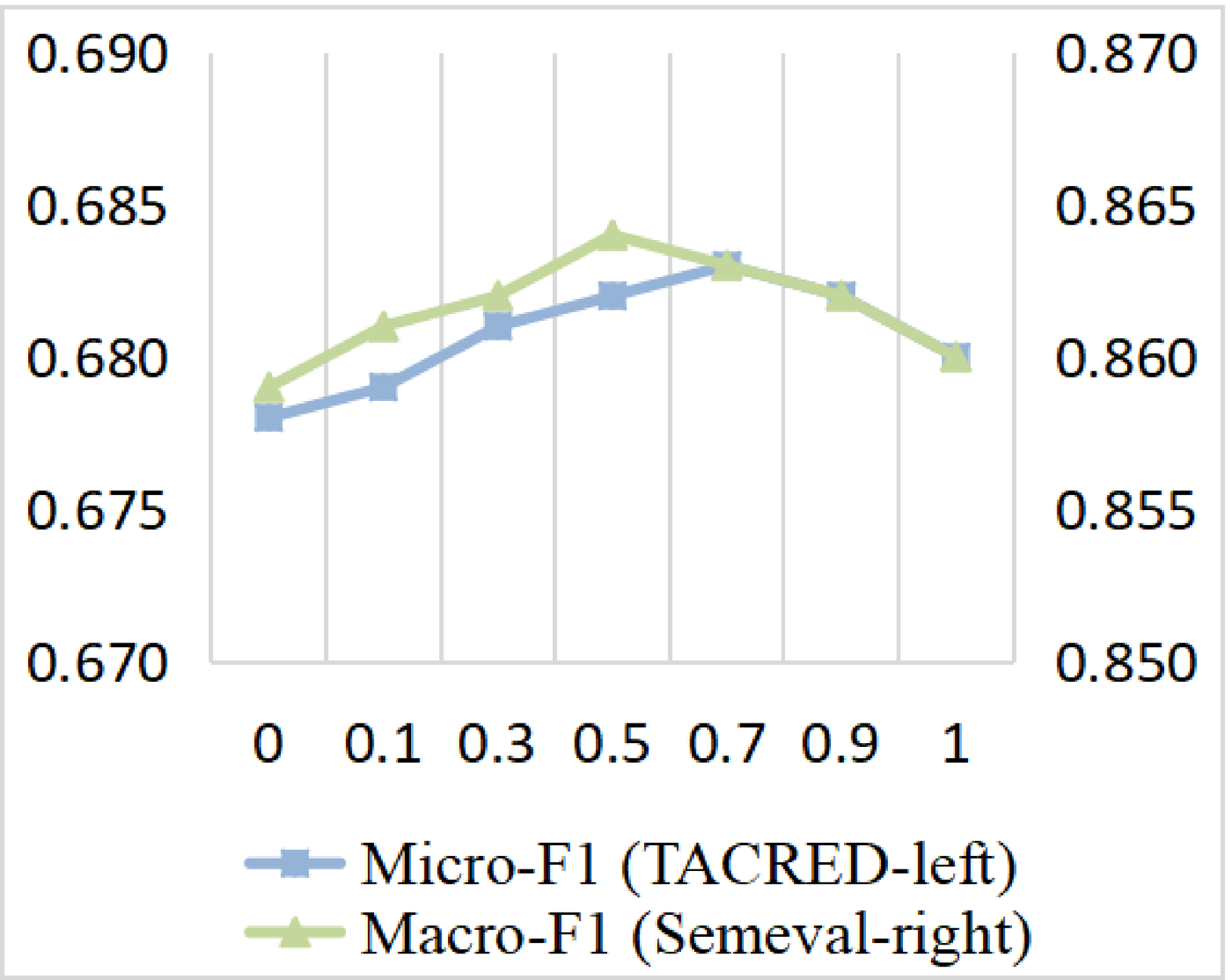}
			\subcaption{$\alpha$}
		\end{minipage}
		\begin{minipage}[t]{0.49\linewidth}
			\centering
			\includegraphics[width=1\textwidth]{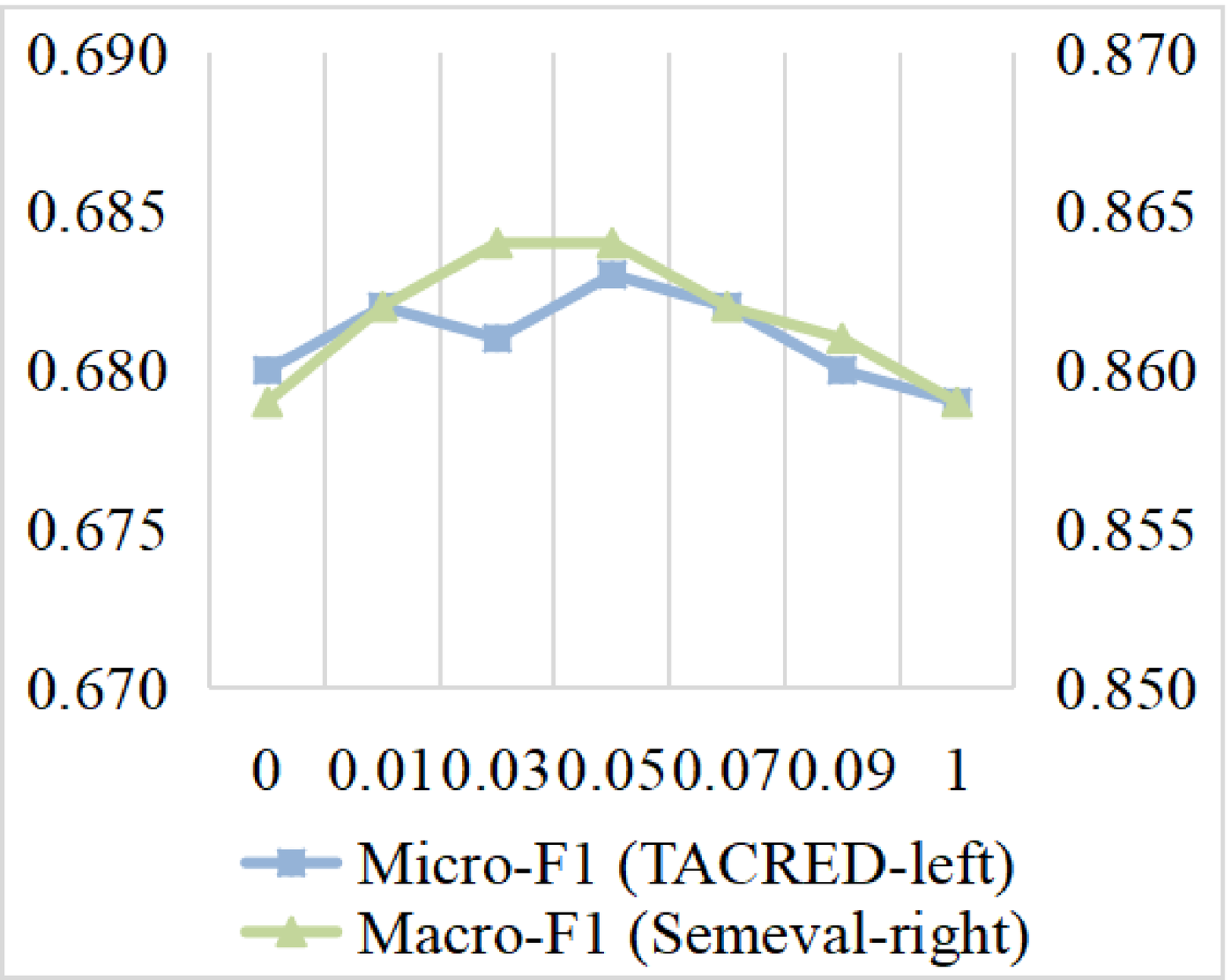}
			\subcaption{$\beta$}
		\end{minipage}
		\caption{Impacts of two hyper-parameters $\alpha$ and $\beta$ on TACRED and Semeval.}	
		\label{figsen:para}
	\end{figure}

	Our model has two key  hyper-parameters $\alpha$ and $\beta$ in Eq.~\ref{loss_all}, which determine the weights of \textit{loss}$^{\textit{rel}}$ and \textit{loss}$^{\textit{rep}}$, respectively. This section examines the impacts of $\alpha$ and $\beta$ on PubMed, TACRED, and Semeval dataset with GloVe embeddings. Because the trend of ``Single'' is the same with that of ``Cross'' in binary-class relation extraction on PubMed dataset, we only report the accuracy for ``Cross''. The results are shown in Figure~\ref{fig:para} and Figure~\ref{figsen:para}, respectively.
	
	In general, the curves for $\alpha$ and $\beta$ have similar trends in that they reach a peak first and then gradually decrease.
	Remember that \textit{loss}$^{\textit{rel}}$ and \textit{loss}$^{\textit{rep}}$ are used to guide the relation-level and feature-level metric learning. One is for relation label between entities for relation extraction and the other for entity representation.
	If they are set too small, the model cannot get the guidance from the metric learning module. If they are set too large, the training  process inclines towards the metric learning subtask and thus the main classification performance decreases.
	
	\begin{table}[!htbp]
		\small
		\centering	
		\setlength{\tabcolsep}{1mm}
		\caption{The model size (M) of our model and that of typical baselines on PubMed, TACRED, and Semeval.}
		\vspace{-2mm}
		\begin{tabular}{lcccccc} \\
			\toprule[1pt]
			\multirow{4}[0]{*}{\textbf{Model}}& \multicolumn{4}{c}{\textbf{PubMed}} & \multirow{4}[0]{*}{\textbf{TACRED}}  & \multirow{4}[0]{*}{\textbf{Semeval}}\\
			\cmidrule(r){2-5}
			&\multicolumn{2}{c}{\textbf{Binary-class}} & \multicolumn{2}{c}{\textbf{Multi-class}}  \\
			\cmidrule(r){2-3} \cmidrule(r){4-5}
			&\textbf{T} & \textbf{B} & \textbf{T} & \textbf{B}\\
			\cmidrule(r){1-5} \cmidrule(r){6-6} \cmidrule(r){7-7}	
			PA-LSTM~\cite{ZhangZCAM17}&	1.77&	1.83 &	1.77	&1.84 &10.8&	9.9\\
			AGGCN~\cite{GuoZL19}&	1.74 &	1.64&	1.74&	1.64&	20.4&	11.0\\
			C-GCN-MG~\cite{MandyaBC20}&	5.60&	5.54&	5.64&	5.57&	96.4&	48.7\\
			EDCRE~\cite{LeeSOLSL20}&	5.87&	5.86&	5.87	&5.89&	98.5&	54.8\\
			\cmidrule(r){1-5} \cmidrule(r){6-6} \cmidrule(r){7-7}
			Our model&	4.18&	4.76&	4.81&	4.32&	84.8&	40.64\\
			\midrule[1pt]		
		\end{tabular}%
		\vspace{-2mm}
		\label{m_size}%
	\end{table}%

	\subsection{Computational Cost Analysis}
	We compare the model size (the number of parameters) of our model to other four typical baselines with GloVe embeddings on PubMed, TACRED, and Semeval in Table~\ref{m_size}. We use an open source PyTorch implementation\footnote{\href{https://github.com/songyouwei/ABSA-PyTorch}{https://github.com/songyouwei/ABSA-PyTorch}.} to measure the model size for these methods.
	
	It can be seen that using the same dimension of hidden states, PA-LSTM~\cite{ZhangZCAM17} and AGGCN~\cite{GuoZL19} have a lower model size compared with other baselines.
	Our model is in the middle in terms of the model size but it can produce the best classification results.

	\section{Conclusion}
	In this paper, we propose a novel multi-scale feature and metric learning framework. Our contributions are three-fold. Firstly, we design a multi-scale convolution neural network (MSCNN) with dilated horizontal and vertical convolution to extract non-successive lexical patterns from the word sequence. Secondly, we design a multi-scale graph convolution network (MSGCN) to enlarge the receptive field of the word along different syntactic paths. Thirdly, we present a feature-level and relation level metric learning scheme to further refine the representations.
	
	We conduct extensive experiments on three datasets for cross sentence $n$-ary and sentence-level RE tasks. The results prove that our model achieves a new state-of-the-art performance with both the static and contextualized word embeddings.

	
	\section*{Acknowledgment}
	This work has been supported in part by the NSFC Projects (61572376, U1811263, 62032016, 61972291).

	\bibliographystyle{IEEEtran}
	\bibliography{ref_s}
	
	%
	

	
	

\end{document}